%% file: 0-main-health.tex
\documentclass[acmlarge,screen,prologue,table]{acmart}
\usepackage{graphicx}
\usepackage{algorithm}
\usepackage{algpseudocode}
\usepackage{subfigure}
\usepackage{booktabs}
\usepackage{multirow}
\usepackage{multicol}
\usepackage{adjustbox}
\usepackage{pdfpages}
\AtBeginDocument{%
  }

\setcopyright{acmlicensed}
\copyrightyear{2018}
\acmYear{2018}
\acmDOI{XXXXXXX.XXXXXXX}

\acmConference[Conference acronym 'XX]{Make sure to enter the correct
  conference title from your rights confirmation emai}{June 03--05,
  2018}{Woodstock, NY}
\acmISBN{978-1-4503-XXXX-X/18/06}

\newcommand{\rev}[1]{{\textcolor{black}{{#1}}}}
\newcommand{\newchange}[1]{{\textcolor{black}{#1}}}

\begin{document}

% \title[Discovering Causes and Effect Modifiers in Practice]{Discovering Causes and Effect Modifiers in Practice: Applications to Repeated Emergency Room Visits and Hospital Readmissions}
%\title[Ensemble Causal Discovery for Identifying Causal Factors of Patient Outcomes]{\newchange{Ensemble Causal Discovery for Identifying Causal Factors of Patient Outcomes: Applications to Healthcare Overutilization}}
%\title{Heterogeneous Causal Discovery of Repeated Undesirable Health Outcomes: Actionable Insights into Repeated Healthcare Utilization}
\title{Heterogeneous Causal Discovery of Repeated Undesirable Health Outcomes}
\author{Shishir Adhikari}
\email{shishir.adhikari@mssm.edu}
\affiliation{%
  \department{Department of Computer Science}
  \institution{University of Illinois Chicago}
  \city{Chicago}
  \state{IL}
  \country{USA}
}
\affiliation{%
  \department{Division of Data-Driven and Digital Medicine}
  \institution{Icahn School of Medicine at Mount Sinai}
  \city{New York}
  \state{NY}
  \country{USA}
}
\authornote{Revision, including clinical and external cohort validation, was performed while being affiliated with ISMMS.}

\author{Guido Muscioni}
\affiliation{%
  \institution{Elevance Health (formerly, Anthem Inc.)}
  \city{Chicago}
  \state{IL}
  \country{USA}
}
\authornote{This work was performed while the authors were affiliated with Elevance Health.}

\author{Mark Shapiro}
\affiliation{%
  \institution{Elevance Health (formerly, Anthem Inc.)}
  \city{Chicago}
  \state{IL}
  \country{USA}
}
\authornotemark[2] % Reuses the footnote marker from Guido Muscioni

\author{Plamen Petrov}
\affiliation{%
  \institution{Elevance Health (formerly, Anthem Inc.)}
  \city{Chicago}
  \state{IL}
  \country{USA}
}
\authornotemark[2] % Reuses the footnote marker from Guido Muscioni

\author{Elena Zheleva}
\email{ezheleva@uic.edu}
\affiliation{%
  \department{Department of Computer Science}
  \institution{University of Illinois Chicago}
  \city{Chicago}
  \state{IL}
  \country{USA}
}


% \author{Shishir Adhikari\thanks{University of Illinois at Chicago (\{sadhik9, ezheleva\}@uic.edu)}
% \and Guido Muscioni\thanks{Anthem Inc (guido.muscioni@anthem.com)}
% \and Mark Shapiro\thanks{Carelon (mark.shapiro@carelon.com)}
% \and Plamen Petrov\thanks{Hydrogen Health (plamen.petrov@khealth.com)}
% \and Elena Zheleva\footnotemark[1]}

% \author{Shishir Adhikari}
% \email{sadhik9@uic.edu}
% \affiliation{
% \institution{University of Illinois Chicago}
% \country{USA}
% }
% \author{Guido Muscioni}
% \email{guido.muscioni@anthem.com}
% \affiliation{
% \institution{Anthem Inc.}
% \country{USA}
% }
% \author{Mark Shapiro}
% \email{mark.shapiro@carelon.com}
% \affiliation{
% \institution{Carelon}
% \country{USA}
% }
% \author{Plamen Petrov}
% \email{plamen.petrov@khealth.com}
% \affiliation{
% \institution{Hydrogen Health}
% \country{USA}
% }
% \author{Elena Zheleva}
% \email{ezheleva@uic.edu}
% \affiliation{
% \institution{University of Illinois Chicago}
% \country{USA}
% }

\setcopyright{none}
\settopmatter{printacmref=false}
\renewcommand\footnotetextcopyrightpermission[1]{}

\begin{abstract}
% This work aims to make causal discovery more practical by considering multiple assumptions and identifying heterogeneous effects. We formulate the problem of discovering causes and effect modifiers of an outcome, where effect modifiers are contexts (e.g., age groups) with heterogeneous causal effects. Then, we present a novel, end-to-end framework that incorporates an ensemble of causal discovery algorithms and estimation of heterogeneous effects to discover causes and effect modifiers that trigger or inhibit the outcome.
% We demonstrate that the ensemble approach improves robustness by enhancing recall of causal factors while maintaining precision. Our study examines the causes of repeat emergency room visits for diabetic patients and hospital readmissions for ICU patients. Our framework generates causal hypotheses consistent with existing literature and can help practitioners identify potential interventions and patient subpopulations to focus on.
Understanding the factors that trigger or prevent undesirable health outcomes across patient subpopulations is essential for designing targeted interventions. While randomized controlled trials and expert-led patient interviews are standard methods for identifying these factors, they can be time-consuming or infeasible. Causal discovery offers an alternative to conventional approaches by generating cause-and-effect hypotheses from observational data, yet its practical utility is limited by strong or untestable assumptions. This work presents a novel, end-to-end framework that uniquely integrates an ensemble of causal structure learning (CSL) algorithms with heterogeneous causal effect estimation. By aggregating results across multiple algorithms, the framework identifies robust causal relationships that persist under different modeling assumptions while simultaneously revealing how these effects vary across specific patient contexts. The proposed heterogeneous causal discovery framework improves robustness and provides practitioners with a prioritized set of actionable, clinically interpretable hypotheses. We demonstrate the framework's effectiveness through two large-scale healthcare applications: identifying drivers and inhibitors of repeat emergency department visits among diabetic patients and hospital readmissions among ICU patients, using insurance claims and electronic health record datasets. Our results, across both settings, identify chronic disease management and care coordination as key interventions, while revealing that intervention effectiveness depends on specific patient-level modifiers. To address the challenge of missing causal ground truth in real-world data, we employ a multi-layered validation strategy. First, simulation-based experiments to evaluate ground-truth recovery reveal that the ensemble approach improves robustness by enhancing recall of causal factors while maintaining precision, driven by high consensus among algorithms on true causes (>90\% agreement) and low consensus on false causes (about 34\% agreement). Second, alignment with clinical literature verifies consistency with established findings. Third, validation by expert clinicians also shows greater consensus among the ensemble's algorithms for true causal relationships than for non-existent ones. %a reliable single model from 65.4\% to 69.6\%. 
Finally, we demonstrate the portability of our framework in modern healthcare systems using an external dataset and discuss how the framework can be useful to clinical investigators.

\end{abstract}

\maketitle

\input{1-introduction}
\input{2-related-work}

\input{3-preliminaries}
\input{3-problem-setup}
\input{4-assumption}
\input{5-methodology}
\input{6-experiments}

\input{7-discussion}

\input{8-conclusion}
\bibliographystyle{ACM-Reference-Format}
\bibliography{references}
\input{9-appendix}
% Insert the entire PDF document
\includepdf[pages=-]{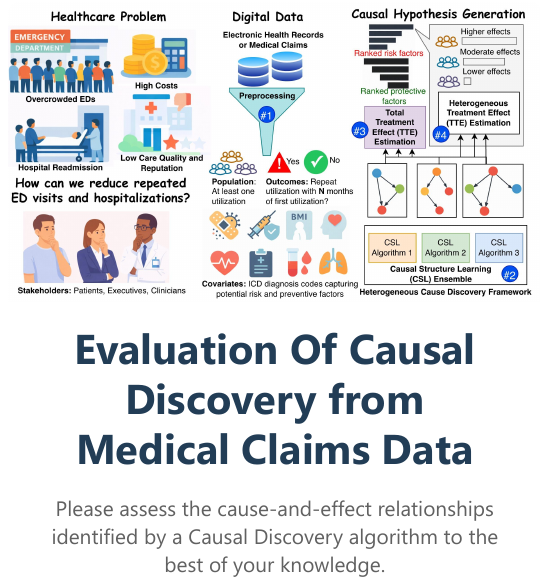}
\end{document}

%% file: 1-introduction.tex
\section{Introduction}

A major challenge in modern healthcare systems is avoidable repeated healthcare utilization, such as repeated
emergency department visits and hospital readmissions~\cite{committee-nap07,cms-web21,nehi-online10}. 
%Avoidable healthcare utilization, such as repeated emergency department visits and hospital readmissions, has been a significant challenge for healthcare systems
Such events often signal gaps in disease management, barriers to accessing routine outpatient care, or insufficient care coordination following discharge~\cite{coleman-aim06,van-bmj23,balasubramanian-jamanet25,malone-jhppl98,lacalle-aem10,hudon-hsr17}. 
These events are costly for healthcare systems and disruptive for patients, and reducing avoidable utilization has become a major priority for clinicians, insurers, and policymakers. Emergency departments (EDs), often referred to as emergency rooms (ERs), play a critical role in providing rapid care for acute medical needs %\rev{In the United States alone, across $107.4$ million treat‑and‑release ED encounters in $2021$, aggregate costs summed to $\$80.3$ billion~\cite{roemer-hucpsp24}.} 
but many healthcare systems face persistent ED overcrowding and operational strain due to repeated (avoidable) ED utilization~\citep{trzeciak-emj03,salway-rmclc17}. %. A substantial fraction of ED encounters occur among patients who return repeatedly, and such 
%Repeated ED visits are often associated with chronic disease burden, poor care coordination, or barriers to accessing routine outpatient care~\citep{malone-jhppl98,lacalle-aem10,hudon-hsr17}. 
Hospital readmissions represent a closely related challenge. Patients who return to the hospital shortly after discharge frequently experience unresolved complications or inadequate post-discharge support~\cite{network-psnet19,balasubramanian-jamanet25}. Because readmissions can reflect both clinical and systemic shortcomings, they have become an important quality metric in healthcare systems. In the United States, policies introduced under the Affordable Care Act publicly report hospital readmission rates and impose financial penalties for excessive readmissions \citep{ppaca-web19,cms-web21}. Despite extensive research on readmission risk factors~\citep{kripalani-arm14}, designing targeted interventions to prevent readmissions remains difficult because of complex interactions among patient characteristics, clinical conditions, and care processes.

Understanding the causal drivers of repeated healthcare utilization is therefore essential for designing effective interventions. Most existing studies use predictive modeling techniques, including logistic regression and modern machine learning methods~\cite{liu-ijcm25,sheikh-wjem19,glans-bmc20,wang-bmcem25,yan-jem17}, that are useful for risk stratification. 
However, these approaches primarily capture statistical associations rather than causal relationships. Consequently, they provide limited guidance for intervention design because variables that improve prediction accuracy may not correspond to factors that can be modified to change outcomes. While randomized controlled trials (RCTs) are the gold standard for establishing causal relationships, they are costly, time-consuming, and often impractical for studying interventions in healthcare utilization. In addition, the large number of interacting variables involved in healthcare utilization, such as chronic disease progression, care coordination practices, and patterns of outpatient engagement, makes
it difficult to design controlled experiments that test all plausible mechanisms. %In addition, the large number of interacting variables involved in healthcare utilization, such as chronic disease progression, care coordination practices, and patterns of outpatient engagement, should be accounted for when developing relevant causal hypotheses and designing controlled experiments. 
Similarly, expert-led patient interviews~\cite{cooksley-am15,smeekes-egm23}, sometimes used to understand causal factors associated with repeated healthcare utilization, may not be able to capture these complex interactions.

Observational healthcare data, such as administrative claims and electronic health records, offer an alternative source of evidence. They capture a rich history of diagnoses, procedures, medications, and healthcare encounters that can be used to generate data-driven hypotheses and evidence. However, extracting actionable insights from such data remains challenging because healthcare events are highly interconnected and the relevant causal relationships are rarely known in advance. Causal discovery methods aim to address this challenge by learning a graph of potential cause–and–effect relationships directly from observational data \citep{spirtes-book00}. Unlike causal inference from observational data~\cite{rubin-jep74,pearl-book09,kunzel-pnas19,stuart-stats10,dimick-jama14}, causal discovery approaches do not require researchers to pre-specify treatment and confounders. %structure that explains the observed dependencies among variables. 
In principle, these methods can generate causal hypotheses about clinical factors or care patterns that influence outcomes such as repeated ED visits or hospital readmissions.
Applying causal discovery in healthcare settings presents important methodological challenges. Different causal discovery algorithms rely on distinct assumptions and search strategies and may produce different causal graphs when applied to the same dataset. This variability raises concerns about the stability and reliability of discovered relationships, particularly in high-dimensional healthcare data where many variables are correlated.

% \begin{figure}
%     \centering
%     \includegraphics[width=0.95\linewidth]{images/ACMHealth (2).jpg}
%     \caption{Caption}
%     \label{fig:framework_overview}
% \end{figure}

In this work, we introduce a framework for identifying robust and actionable drivers of repeated healthcare utilization from observational data. The proposed approach combines two complementary components that are typically studied separately. First, we apply an ensemble of causal structure learning algorithms to discover candidate causal relationships associated with repeated healthcare utilization. Aggregating results across multiple algorithms allows us to identify relationships that are consistently supported under different modeling assumptions, improving robustness relative to relying on a single discovery method. Second, we quantify the magnitude and variability of these relationships by estimating causal effects and examining how these effects change across patient contexts. This heterogeneous effect analysis enables the identification of factors whose influence on the outcome depends on patient characteristics or care pathways. By integrating causal discovery with heterogeneous effect estimation, the framework produces a ranked set of consistent drivers that both (i) appear repeatedly across causal discovery algorithms and (ii) demonstrate meaningful causal effects in specific contexts. The resulting output is not simply a causal graph but a prioritized set of clinically interpretable hypotheses about factors that may increase or decrease the risk of repeated healthcare utilization.
% Figure~\ref{fig:framework_overview} illustrates the overall workflow of the proposed framework, including the formulation of repeated utilization outcomes, the ensemble causal discovery process, the estimation of heterogeneous causal effects, and the recommendation of actionable hypotheses that can subsequently be evaluated through clinical investigation and external validation.

We evaluate the proposed framework through three healthcare applications. The first application examines repeat ED visits among patients with clinical or subclinical diabetes using a large administrative claims dataset comprising $461{,}754$ records from a major health insurer. Although claims data are primarily collected for billing purposes, they capture longitudinal information about diagnoses, procedures, and healthcare encounters across multiple providers, making them a valuable resource for studying patterns of healthcare utilization.
The second application investigates hospital readmissions using the Medical Information Mart for Intensive Care (MIMIC-IV) dataset \citep{johnson-mimiciv21}, which contains detailed electronic health record data from patients under critical care. We analyze $335{,}378$ admission records to identify clinical and care-related factors associated with readmission risk following hospitalization. 
\rev{To complement our primary analysis in vulnerable, high‑risk populations, we apply our framework to a third dataset, EHRSHOT, comprising electronic health records of $6{,}739$ general, mixed‑risk patients from Stanford Health~\cite{wornow-neurips23}. We demonstrate the portability of our framework and investigate whether the causal drivers of repeat ED visits and readmissions are localized to specific risk profiles or persist across general populations. Our a priori hypothesis is that the pattern of insights will diverge, reflecting differences in baseline risk, multimorbidity profiles, care access, and service use between high‑risk patients and the general population.}

Because causal ground truth is rarely available in real-world healthcare datasets, evaluating causal discovery methods is inherently challenging. We address this issue using complementary strategies. First, we examine whether the discovered cause-and-effect hypotheses are consistent with prior clinical and health services research findings. Second, we conduct experiments on simulated datasets where the true causal structure is known, enabling controlled evaluation of the framework’s ability to recover ground-truth relationships. \rev{Third, we recruit four clinicians to independently review a subset of the cause-and-effect relationships discovered from the administrative claims dataset. Finally, we discuss the utility of generated causal hypotheses for clinical investigators, including prioritizing context‑specific intervention options, specifying eligibility criteria, and designing prospective studies optimized for likely‑to‑benefit populations.}

% Beyond identifying candidate drivers and preventive factors of repeated utilization, we also demonstrate how the resulting causal hypotheses can support clinical investigation. Specifically, we illustrate how clinicians can use insights to formulate and evaluate testable hypotheses about care pathways. As a proof of concept, we select a hypothesis from the causal analysis that is not well established in the existing literature and assess it through external validation in the EHR-SHOT dataset using a target trial emulation framework that includes cohort construction and quasi-experimental causal inference. This process demonstrates how causal discovery can serve as a starting point for clinically grounded investigation rather than as a substitute for rigorous causal evaluation. 

The contributions of our work are as follows: 
% First, from a research problem perspective, we investigate a novel problem of jointly discovering causes and effect modifiers that trigger or prevent a healthcare outcome, given uncertainty about the underlying data-generating mechanism. Second, we present a novel computational framework that uniquely integrates an ensemble of causal structure learning algorithms with heterogeneous causal effect estimation to robustly generate causal hypotheses across different modeling assumptions and diverse patient contexts. Third, we conduct a robust empirical evaluation using large-scale insurance claims and electronic health record datasets, validating the framework’s efficacy to identify clinically relevant factors associated with repeated utilization through alignment with the clinical literature, simulation-based ground-truth recovery, and expert clinician review. Fourth, we demonstrate how novel causal hypotheses derived from the analysis can be further validated through target trial emulation and quasi-experimental causal inference methods using comprehensive electronic health records.

\begin{itemize}

\item \textbf{Research problem.} We investigate a novel problem of jointly discovering causes and effect modifiers that trigger or prevent an outcome, given uncertainty about the underlying data-generating mechanism.

\item \textbf{Methodological contribution.} We present a novel computational framework that uniquely integrates an ensemble of causal structure learning algorithms with heterogeneous causal effect estimation to robustly generate causal hypotheses across different modeling assumptions and diverse population characteristics.

\item \rev{\textbf{Empirical evaluation.} We conduct a robust empirical evaluation using large-scale insurance claims and electronic health record datasets, validating the framework’s efficacy to identify clinically relevant factors associated with repeated utilization through alignment with the clinical literature, simulation-based ground-truth recovery, and expert clinician review.}

\item \rev{\textbf{Real-world impact.} We develop a framework for actionable insights into repeated healthcare utilization. We demonstrate how automatically generated causal hypotheses can be further corroborated to assess generalization using comprehensive external electronic health records and discuss how the framework can be particularly useful for clinical investigators in formulating and evaluating testable hypotheses about care pathways.}

\end{itemize}

%% file: 2-related-work.tex
% \vspace{-1em}

\section{Related Work}\label{sec:relatedwork}

\newchange{\textbf{Analysis of repeat ER visits and hospital readmissions}. There are multiple studies on identifying the factors contributing to repeat ER visits and readmissions, including meta-analyses~\cite{soril-plos15,sheikh-wjem19,leppin-jama14}, retrospective observational studies~\cite{hudon-hsr17,glans-bmc20}, prospective cohort studies~\cite{ross-aem10,noori-ijhpm14}, and PRISMA-based root cause analysis~\cite{van-book97,driesen-bmcg20,cooksley-am15}. The PRISMA (Prevention and Recovery Information System for Monitoring and Analysis) method involves trained investigators manually constructing a causal tree after systematic interviews with the patients. 
There is a growing interest in identifying and analyzing heterogeneous subpopulations associated with recurrent ER visits and readmissions by utilizing observational and interventional data~\cite{soril-plos15,birmingham-ajem20,penney-bmc18}.
Our approach uses retrospective observational data to automatically discover potential causes and sources of heterogeneity, assisting investigators in identifying and designing interventions.}

% \fixme{TODO: Fix the issue with command citet}
\newchange{\textbf{Causal discovery in healthcare domains}. Causal discovery approaches have been employed in healthcare domains, utilizing different background information. \citet{wang-ieee20} have used local causal structure learning with prior domain knowledge to discover causal relationships between Type 2 Diabetes and Bone Mineral Density (BMD)- related conditions. Another approach ~\cite{hu-acmhealthcare23} has used prior knowledge extracted from multiple Authoritative Medical Ontologies (AMOs) to orient edges learned by a causal Bayesian learning algorithm. \citet{nordon-aaai19} used causal relations extracted from medical literature to construct a causal graph and electronic medical record (EMR) data to prune the edges with no correlation. Our approach uses an ensemble of causal discovery algorithms, assuming that the patient outcome (e.g., readmission) is never a parent of other contexts (e.g., diagnosis codes) due to the temporal order of occurrence.}

\textbf{Causal structure learning approaches}. Causal structure learning (CSL) algorithms can be classified into five types according to their approach: (1) constraint-based, (2) score-based, (3) hybrid, (4) functional causal model-based, and (5) continuous optimization-based. The idea behind CSL is to exploit conditional independencies (CIs) in the data distribution to infer the structure of the causal graphical model. The notion of \textit{d-separation}~\cite{pearl-book88,pearl-book09} is used to read off all CI that hold for any data distribution that is generated by the mechanism described by a graphical model. The same conditional independence relation can be satisfied by multiple causal models within a Markov equivalence class, and CSL typically concerns learning a Markov equivalence class. Constraint-Based algorithms, such as PC~\cite{spirtes-book00,colombo-jmlr14} and Fast Causal Inference (FCI) \cite{spirtes-book00}, use conditional independencies in the data as constraints to estimate an equivalence class of underlying SCM. Score-based algorithms, such as Greedy Equivalence Search (GES)~\cite{chickering-jmlr02} and Fast GES~\cite{ramsey-ijdsa17}, view causal structure learning as the problem of fitting a CPDAG to the data using a relevant score function that reflects how well the CPDAG captures the conditional independencies in the data. Hybrid algorithms combine both constraint-based and score-based approaches. Max-Min Hill Climbing (MMHC)~\cite{tsamardinos-ml06} uses a constraint-based Max-Min Parents and Children (MMPC) local discovery algorithm for skeleton determination and a score-based Hill-Climbing (HC)~\cite{scutari-jss10} for edge orientation. The functional causal models~\cite{glymour-fg19} use SCM with additional assumptions on the distribution of \rev{endogenous variables and exogenous noise to identify causal orientation from DAGs in} the same equivalence class.

Recently, methods using continuous optimization~\cite{zheng-nips18,zheng-aistats20} are popular for extracting a directed acyclic graph (DAG) structure from observational data. Although these gradient-based techniques are scalable, the output DAG lacks a causal interpretation and these methods are less reliable with real-world data~\cite{kaiser-springer22, olko-icml25}. 
% It is demonstrated that these models lack scale-invariance and perform poorly on real-world data~\cite{kaiser-springer22}.
The CSL algorithms and their assumptions with empirical performances have been reviewed with a continuous data~\cite{heinze-arsa18}. Previous work on causal effect estimation combining CSL and covariate adjustment focused on identifying a multi-set of causal effects for the equivalence class of a single DAG~\cite{maathuis-astat09,perkovic-jmlr17}. 
Here, we study the behavior of CSL algorithms and their ensemble with categorical non-linear data for discovering causes and effect modifiers of a sink outcome node.
There are approaches~\cite{cai-tnnls22,gong-arxiv23} for causal discovery using event sequences that may seem more relevant for medical claims data or electronic health records. These algorithms rely on the crucial assumption that a cause precedes an effect temporally. The discussed section presents our rationale for relying on the summary graph and for aggregating only the historical presence or absence of diagnosis codes for a patient, which leads to the observed outcome.

%% file: 3-preliminaries.tex
\section{Preliminaries}
\subsection{Structural Causal Model (SCM)}
% \vspace{-0.5em}
A SCM {$\mathcal{M}(\mathbf{V}, \mathbf{U}, \mathbf{f})$} models the underlying data-generating mechanism with a set of exogenous variables {$\mathbf{U}$}, endogenous variables {$\mathbf{V}$}, and functions {$\mathbf{f}$}.
% The exogenous variables {  $\mathbf{U}$} are external to the causal model and considered errors or disturbances. 
A variable is a function of its known direct causes and unknown disturbances.
The causal relationship among the variables in an SCM is represented using a directed acyclic graph (DAG) {$G(\mathbf{V}, \mathbf{E})$} where {$\mathbf{V}$} and {$\mathbf{E}$} are a set of vertices and edges, respectively, and each vertex has incoming edges from its direct causes. 
{$V_i \rightarrow V_j$} denotes an edge from the parent {$V_i$} to the child {$V_j$}.
Two vertices are adjacent if there is an edge between them.
A directed path is a sequence of nodes obtained by following the direction of the edges. A graph is directed acyclic if there are no directed paths that contain a repeated node. The nodes preceding the tail node of the directed path are the ancestors of the tail node. The nodes following the head node of the directed path are descendants of the head node.
The symbols {$pa(V_i, G)$}, {$anc(V_i, G)$}, and {$des(V_i, G)$} indicate the sets of parents, ancestors, and descendants of {$V_i$} for graph {$G$}, respectively. Given a ground truth $G$, the causes of an outcome $Y$ are its ancestors.%, i.e., $anc(Y, 

\subsection{Effect modifier (EM)} 
Although the presence of effect modification is determined by the estimation of CATE (Eq. \ref{eq:hte}), the graphical structure $G$ can still inform about potential interactions that result in effect modification. \citet{vanderweele-epi07} use DAGs to categorize four types of potential effect modifiers: direct, indirect, common cause, and proxy. All effect modifiers are non-descendants of the treatment and outcome.
Direct EMs are the parents of the outcome except the treatment. Indirect, common cause, and proxy EMs are related to the direct EM by ancestor, confounding, and descendant relationships, respectively. In general, the potential interacting variables for a given treatment and outcome in a $G$ include the non-descendants of the treatment node and either (1) the parents of the outcome or (2) the parents of mediator nodes in the directed path from treatment to outcome~\cite{vanderweele-epi07}.
% \citet{vanderweele-epi07} have studied identification and categorization of effect modifiers in causal graphs.

\subsection{Conditional Average Treatment Effect (CATE) Identification}  The cause-and-effect hypotheses encoded in $G$ can be used to identify causal effects from observation data using do-calculus~\cite{pearl-book09}.
Here, we show backdoor adjustment~\citep{pearl-book09} for the identification of causal effects. Let $\mathbf{W}$ be an adjustment set such that $\mathbf{W} \cup \mathbf{Z}$ satisfies the backdoor criterion for treatment $X$, outcome $Y$, and contexts $\mathbf{Z}$. The interventional distributions {$E[Y|do(X=x), \mathbf{Z}=\phi]$} and {$E[Y|do(X=x), \mathbf{Z}=\mathbf{z}]$} can be estimated, respectively, in terms of observational distribution with backdoor adjustment as:
\begin{equation}\label{eq:cateobs}
\begin{split}
    {E[Y|do(X=x), \mathbf{Z}=\phi] = \textstyle\sum_{\mathbf{w}}{E[Y|X=x, \mathbf{W}=\mathbf{w}]P(\mathbf{W}=\mathbf{w})}}, \text{and}\\
    {E[Y|do(X=x), \mathbf{Z}=\mathbf{z}] = \textstyle\sum_{\mathbf{w}}{E[Y|X=x, \mathbf{W}=\mathbf{w}, \mathbf{Z}=\mathbf{z}]P(\mathbf{W}=\mathbf{w}|\mathbf{Z}=\mathbf{z})}}.
\end{split}
\end{equation}
% We can estimate $X_{CATE}(\mathbf{Z}=\mathbf{z})$ by substituting two expectation terms in Equation \ref{eq:ate} with Equation \ref{eq:cateobs}. 
To satisfy backdoor criteria, $\mathbf{W}$ should not be descendants of $X$ and $\mathbf{W} \cup \mathbf{Z}$ should block all backdoor paths from $X$ to $Y.$  
These estimands can be substituted in Equation \ref{eq:ate} to estimate CATE.

\subsection{Causal Structure Learning (CSL)}

% \fixme{TODO: Add section on assumptions like causal sufficiency, faithfulness, causal minimality, etc.} \\

\newchange{The idea for CSL is to utilize conditional independencies (CI) in the data distribution to infer the structure of the causal graphical model.
The joint distribution $P$ described by $G(\mathbf{V}, \mathbf{E})$ factorizes to the product of the conditional probability of each random variable given its parents according to the causal Markov assumption, i.e.,
    {\small $P(V_1,...,V_n) = \prod_{i=1}^{n} P(V_i | pa(V_i, G)).$}
The notion of \textit{d-separation}~\cite{pearl-book88,pearl-book09} is used to read off all CI that hold for any data distribution that is generated by the mechanism described by a graphical model. The same conditional independence relation can be satisfied by multiple causal models belonging to a Markov equivalence class, and CSL generally concerns learning a Markov equivalence class. Additional parametric assumptions and background knowledge are needed to identify a causal model within the equivalence class.}
We summarize four categories of causal structure learning algorithms according to their approach: (1) constraint-based, (2) score-based, (3) hybrid, and (4) functional causal model-based. 
Constraint-Based algorithms, such as PC~\cite{spirtes-book00,colombo-jmlr14} and Fast Causal Inference (FCI) \cite{spirtes-book00}, use conditional independencies in the data as constraints to estimate an equivalence class of underlying SCM. The PC algorithm makes assumptions of causal sufficiency and faithfulness for the correctness of edge adjacencies. Then, v-structure discovery followed by Meek orientation rules~\cite{meek-uai95} produces an equivalence class graph, also known as a completed partial directed acyclic graph (CPDAG). FCI is also a constraint-based algorithm that relaxes the causal sufficiency assumption of the PC algorithm. FCI outputs a partial ancestral graph (PAG), a Markov equivalence class of maximal ancestral graph (MAG), to incorporate the hidden confounders using a bidirectional arrow, i.e., $V_i \leftrightarrow V_j$. In a PAG, $V_i \rightarrow V_j$ is interpreted as $V_i$ being an ancestor of $V_j$ and $V_j$ not being an ancestor of $V_i$. Similarly, $V_i$~o$\rightarrow V_j$ indicates either $V_i$ is an ancestor of $V_j$ and/or there is a hidden confounder between $V_i$ and $V_j$, but $V_j$ is not an ancestor of $V_i$. 

%The FCI algorithm and its variants relax the causal sufficiency assumption~\cite{glymour-fg19}.
Score-based algorithms, such as  Greedy Equivalence Search (GES)~\cite{chickering-jmlr02} and Fast GES~\cite{ramsey-ijdsa17}, consider causal structure learning as the problem of fitting a CPDAG to the data according to a relevant score function that relates to how well the CPDAG captures the conditional independencies in the data. FGES's implementation in \texttt{py-causal}~\cite{pycausal-github19} package provides some robustness to the faithfulness assumption. These algorithms assume causal sufficiency. Under the fulfillment of assumptions and infinite sample size with the absence of selection bias, both PC and GES algorithms are shown to be sound and complete.
% For HC and MMHC, we use \texttt{bnlearn}~\cite{scutari-jss10} package. 

Hybrid algorithms combine both constraint-based and score-based approaches. Max-Min Hill Climbing (MMHC)~\cite{tsamardinos-ml06} uses a constraint-based Max-Min Parents and Children (MMPC) local discovery algorithm for skeleton determination and a score-based Hill-Climbing (HC)~\cite{scutari-jss10} for edges orientation. GFCI~\cite{ogarrio-cpgm16} is a hybrid algorithm that relaxes causal sufficiency assumption.

The functional causal models~\cite{glymour-fg19} use SCM with additional assumptions on the distribution of $\mathbf{U}$ and $\mathbf{V}$ to distinguish between different DAGs in the same equivalence class. Linear Non-Gaussian Acyclic Model (LINGAM)~\cite{shimizu-jmlr06} assumes causal sufficiency, linear data generating process, and exogeneous variables with Non-Gaussian distributions of non-zero variance. Although additive noise models for discrete data have been proposed, it is too restrictive for binary variables~\cite{peters-pami11}.
Recently, differentiable causal discovery methods~\cite{zheng-nips18,zheng-aistats20,ng-neurips,bello-neurips22} using continuous optimization with acyclicity constraints are popular for extracting a DAG structure from observational data. While these methods improve scalability, they are less reliable in real-world settings due to their sensitivity to hyperparameters, feature scale, and sample size~\cite{kaiser-springer22,olko-icml25}.
% However, these methods performed poorly on our synthetic data and we excluded them from our analysis.

%Recent directions in causal discovery include ordering-based methods~\cite{pmlr-v162-rolland22a,xu2024ordering,sanchez2023diffusion} and continuous optimization methods~\cite{zheng2018dags,ng2020role,bello2022dagma}. Ordering-based methods aim to recover an ancestral topological ordering between nodes in the causal graph, often through score-based criteria, and then refine the resulting graph by removing spurious edges. Alternatively, methods like NOTEARS \cite{zheng2018dags}, GOLEM \cite{ng2020role}, and DAGMA \cite{bello2022dagma} cast the structure learning problem into a continuous optimization framework with explicit acyclicity constraints. A major benefit of these formulations is their favorable optimization behavior and scalability, particularly due to their compatibility with GPU-based computation.

%% file: 3-problem-setup.tex
% \vspace{-1em}
\section{Problem Setup}\label{sec:problemsetup}
First, we present our data model to represent and abstract patient data used in our study. Then, we formally describe the problem of discovering the causes and effect modifiers of patient outcomes.

\subsection{Data Model}

% \fixme{TODO: Incorporate static patient features as well}\\

\textbf{Event-log data model}. Let {\small $e^j_{i,t} \in \{0, 1\}$} and {\small $o_{i,t} \in \{0, 1\}$} represent an event of interest of type {\small $e^j$} and an undesirable outcome {\small $o$} for unit {\small $i$} at time {\small $t$}. We consider {\small $M$} events {\small $e^j~|~j \in \{1,...,M\}$} that are potentially related to the undesirable outcome {\small $o$}. The binary values $1$ and $0$, respectively, indicate the presence and the absence of the events or a outcome for a given unit and time.  Each independent unit consists of a timeline {\small $E_i \in \{0, 1\}^{(M+1) \times T}$} of time frame {\small $T$} consisting of events and outcomes. 
% We focus on the time-associated events of a unit, such as disease diagnoses, since other attributes, such as socio-demographic information, are often removed in anonymized data. 

\begin{figure}[!t]
    \centering
    \includegraphics[height=0.4\linewidth,keepaspectratio]{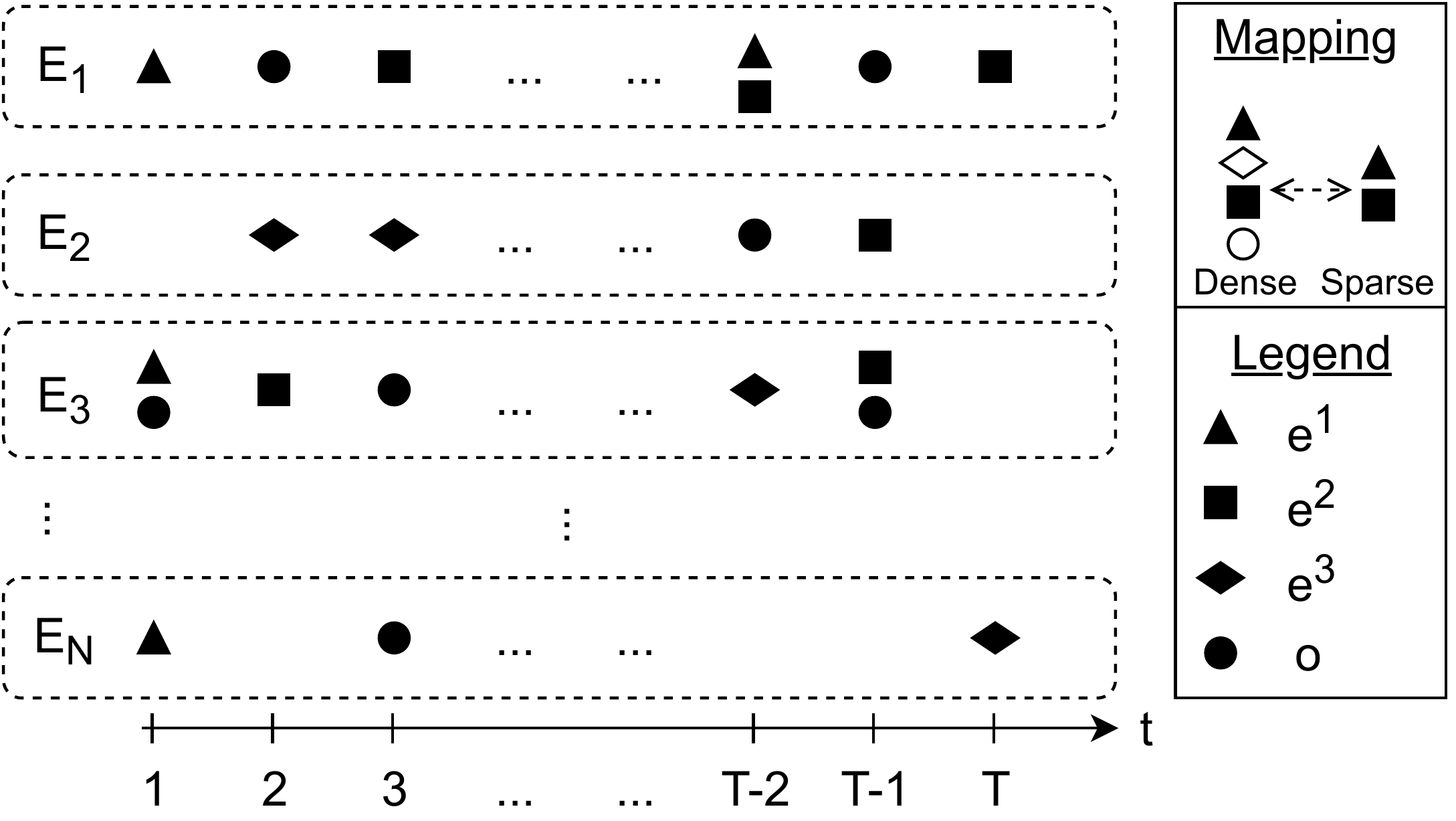}
    \caption{Illustration of data model with timeline $E \in \{0,1\}^{4 \times T}$ for $N$ units with three endogenous events ($e^1, e^2,$ and $e^3$) and an undesirable outcome ($o$). %For visual conciseness, a sparse representation of event vectors is shown.
    % with events \{$e^j|e^j_{i,t}=1$\}.
    }
    \label{fig:data_model}
    \vspace{-1em}

\end{figure}

Figure \ref{fig:data_model} illustrates the data model described above for {\small $N$} units and {\small $M=3$} events with one undesirable outcome. For visual conciseness, a sparse representation of event vectors is shown with events {\small $\{e^j|e^j_{i,t}=1\}$}. The triangle, square, and diamond shapes represent different event types that are potentially related to the undesirable outcome depicted with a circle. The shaded figures indicate the presence of an event. In context to the real-world use case in our study, the undesirable outcome is an ER visit. The events can include diagnoses of diseases, routine office visits, or encounters of immunization. We refer to this data model as \textit{event-log data model}.

%\note{\textbf{Metadata data model}. Metadata data model optionally describes 
Each unit $i$ or event at a timepoint $t$ can have covariates, $\mathbf{X_I}$ or $\mathbf{X_T}$, respectively. %For the sake of brevity, we describe the problem and methodologies with event-log data model. 
For example, for our use case on readmission, we incorporate socio-demographic covariates describing a patient (unit), and temporal contextual covariates describing a hospitalization event.
Next, we define the population of interest and the repeated undesirable outcome using the event-log data model.

% \fixme{TODO: Change ``Repeated undesirable outcome`` to outcome event?} \\

\textbf{Population of interest}. Our population of interest ({\small $\mathbf{P}$}) consists of units with at least one undesirable outcome. Formally, the population is a subset of units such that there exists at least one undesirable outcome in the unit's timeline i.e. {\small$\mathbf{P} \subseteq \mathbf{E}~|~ \forall E_i \in \mathbf{P}, \exists t \in \{1,...,T\}, o_{i,t}=1 $}.

\textbf{Repeated undesirable outcome}. If an undesirable outcome recurs within a specified time interval $\tau$ of the prior undesirable outcome, then the more recent outcome is a repeated undesirable outcome. Let, {\small$t' \in \{t+1,...,T\} \cup \emptyset$} indicate the time point of the undesirable outcome occurring after a reference time point {\small $t$} in which there was an undesirable outcome {\small $o_{i,t}=1$}, and is defined as follows:
{\small
\begin{equation}\label{eq:t'}
    t' := min(p \in \{t+1,...,T\} ~|~ o_{i,p}=1),
\end{equation}}
where $\emptyset$ indicates a null value.

The repeated undesirable outcome {\small $y_{i}(t,t',\tau) \in \{0, 1\}$} is defined by the specified time interval {\small $\tau$}, the reference time point {\small $t$}, and the time point {\small $t'$}:
{\small
\begin{equation}\label{eq:outcome}
    y_{i}(t,t',\tau) :=
\left\{
	\begin{array}{ll}
		0  & \mbox{if } t'=\emptyset \\
		0 & \mbox{if } t' - t > \tau \\
		1  & \mbox{if } t' - t \leq \tau 
		
	\end{array}
\right..
\end{equation}
}
Equation (\ref{eq:outcome}) enables mapping of multiple occurrences of undesirable outcomes at time points {\small $t \in \{1,..,T\}~|~o_{i,t}=1$} to a binary value representing a \textit{repeated} undesirable outcome i.e. {\small $y_{i}({\tau})= 1$} if {\small $\exists{t}, y(t,t',\tau)=1$} and $0$ otherwise.
% \begin{equation}
%  y_{i}({\tau})=
% \left\{
% 	\begin{array}{ll}
% 		1  & \mbox{if } \exists{t}, y(t,t',\tau)=1 \\
% 		0 & \mbox{otherwise}
% 	\end{array}
% \right..
% \end{equation}}
Moving forward, we use the terms ``outcome'' and ``repeated undesirable outcome'' interchangeably.

For example, let us consider the timeline {\small $E_3$} in Figure \ref{fig:data_model}. If the time interval {\small $\tau=2$}, then the outcome is {\small $y_{3}(1,3,2)=1$} because the consecutive undesirable outcomes are within the specified time interval. However, if we consider {\small $\tau=1$} for the timeline {\small $E_3$}, the outcome is {\small $y_{3}(1,3,1)=0$}.
We treat the value of {\small $\tau$} to be domain-specific and available from standard practice or domain experts. For example, the commonly used interval for hospital readmission is $30$ days~\cite{bailey-hcup19}.

\subsection{Discovering causes and effect modifiers}
% \fixme{TODO: Make it journal-friendly (e.g., using illustrations like in the presentation).}
{In this section, we first describe the causes and effect modifiers of an outcome using hypothetical interventional distribution~\citep{pearl-book09} and then formally define our research problems.}

%Let $X^j$ and $Y$ be the random variables for the $j^{th}$ event and the outcome respectively. 
Under stable unit treatment value assumption (SUTVA), a variable $X$ is a cause of outcome $Y$ if the \textit{conditional average treatment effect} (CATE) for any context $\mathbf{Z}=\mathbf{z}$ is non-zero, i.e.,
% \fixme{
{ 
\begin{equation}\label{eq:ate}
% \begin{split}
    \exists{\mathbf{Z}=\mathbf{z}}, X_{CATE}(\mathbf{Z}=\mathbf{z}) :=
    E[Y|do(X=x), \mathbf{Z}=\mathbf{z}] - E[Y|do(X=x'), \mathbf{Z}=\mathbf{z}] \ne 0,
% \end{split}
\end{equation}}
where {  $do(X=x)$} and {  $do(X=x')$} denote interventions to the value of {$X$} with policies {$x$} and {$x'$}, respectively. We assume the context $\mathbf{Z}$ is not affected by the outcome to ensure the soundness of the CATE estimand. %i.e., $\mathbf{Z} \subset \mathbf{V} \wedge \mathbf{Z} \not\subset des(Y,G)$.
% The estimand on Eq. \ref{eq:ate} assumes the variables $\mathbf{Z}$ are observed.
The causal effect in Equation \ref{eq:ate} is measured on a difference scale, but for a binary outcome, the effect may also be measured on a multiplicative scale, such as a risk ratio or an odds ratio. A treatment may have heterogeneous effects on the outcome in different contexts. These contexts are called effect modifiers.
%Next, we discuss how the problem of discovering effect modifiers relates to a SCM, particularly the causal graphical model $G$.
A set of variables {$\mathbf{Z}$} are effect modifiers for treatment {$X$} and outcome {$Y$} if CATE is different for any two contexts {$\mathbf{z}$} and {$\mathbf{z}'$} that are instances of {$\mathbf{Z}$}, i.e.,
{ 
\begin{equation}\label{eq:hte}
    X_{CATE}(\mathbf{Z}=\mathbf{z}) \ne X_{CATE}(\mathbf{Z}=\mathbf{z}').
\end{equation}
}
Next, we formally define two research problems.\\
\textbf{Problem 1} (\textit{Discovering causes of an outcome})\label{def:1} Given a set of variables {$\mathbf{X}$} and the outcome {$Y$}, find a set of variables {$\{X \in \mathbf{X}|\exists{\mathbf{Z}=\mathbf{z}},X_{CATE}(\mathbf{Z}=\mathbf{z}) \ne 0\}$}.\\
\textbf{Problem 2} (\textit{Discovering effect modifiers of a treatment and an outcome})\label{def:2} Given a treatment {$X$} and outcome {$Y$}, find all contexts {$\{\mathbf{z},\mathbf{z}'\} \in \mathbf{Z} \subset \mathbf{X} \setminus \{X,Y\}$} such that Equation \ref{eq:hte} holds.

The magnitude and sign of $X_{CATE}(\mathbf{Z}=\mathbf{z})$ help to distinguish whether causes and effect modifiers are triggering or preventing the outcome. An effect modifier $Z$ for a treatment may or may not be a cause of the outcome. We refer to the causes and causal effect modifiers as \textit{heterogeneous causes} of the outcome and focus on discovering them. Identification of causal effect modifiers can help in effective targeted interventions. 
% For example, vaccination should be prioritized to groups on risk (race) as opposed to proxies (a zip address).

%% file: 4-assumption.tex
\vspace{-0.5em}
\section{Discovering Causes and Effect Modifiers from Observational Data}\label{sec:assumptions}
The research problems described in Section \ref{sec:problemsetup} are defined in terms of interventional distributions. Here, we discuss the opportunities and challenges for addressing the research problems with observational data using the ideas of the structural causal model~\cite{pearl-book09} and causal structure learning~\citep{spirtes-book00}.

% \subsection{Background}
% \textbf{Structural causal model (SCM).} A SCM {$\mathcal{M}(\mathbf{V}, \mathbf{U}, \mathbf{f})$} models the underlying data-generating mechanism with a set of exogenous variables {$\mathbf{U}$}, endogenous variables {$\mathbf{V}$}, and functions {$\mathbf{f}$}.
% % The exogenous variables {  $\mathbf{U}$} are external to the causal model and considered errors or disturbances. 
% A variable is a function of its known direct causes and unknown disturbances.
% The causal relationship among the variables in an SCM is represented using a directed acyclic graph (DAG) {$G(\mathbf{V}, \mathbf{E})$} where {$\mathbf{V}$} and {$\mathbf{E}$} are a set of vertices and edges, respectively, and each vertex has incoming edges from its direct causes. 
% {$V_i \rightarrow V_j$} denotes an edge from the parent {$V_i$} to the child {$V_j$}.
% Two vertices are adjacent if there is an edge between them.
% A directed path is a sequence of nodes obtained following the direction of the edges. A graph is directed acyclic if there are no directed paths with repeated nodes. The nodes preceding the tail node of the directed path are the ancestors of the tail node. The nodes following the head node of the directed path are descendants of the head node.
% The symbols {$pa(V_i, G)$}, {$anc(V_i, G)$}, and {$des(V_i, G)$} indicate the sets of parents, ancestors, and descendants of {$V_i$} for graph {$G$}, respectively. Given a ground truth $G$, the causes of an outcome $Y$ are its ancestors.%, i.e., $anc(Y, G)$.

\textbf{Effect modifier (EM).} 
Although the presence of effect modification is determined by the estimation of CATE (Eq. \ref{eq:hte}), the graphical structure $G$ can still inform about potential interactions that results in effect modification. \citet{vanderweele-epi07} use DAGs to categorize four types of potential effect modifiers: direct, indirect, common cause, and proxy. All effect modifiers are non-descendants of the treatment and outcome.
Direct EMs are the parents of the outcome except the treatment. Indirect, common cause, and proxy EMs are related to the direct EM by ancestor, confounding, and descendant relationships, respectively. In general, the potential interacting variables for a given treatment and outcome in a $G$ include the non-descendants of the treatment node and either (1) the parents of the outcome or (2) the parents of mediator nodes in the directed path from treatment to outcome~\cite{vanderweele-epi07}.
% \citet{vanderweele-epi07} have studied identification and categorization of effect modifiers in causal graphs.

\textbf{CATE identification.}  The cause-and-effect hypotheses encoded in $G$ can be used to identify causal effects from observation data using do-calculus~\cite{pearl-book09}.
Here, we show backdoor adjustment~\citep{pearl-book09} for the identification of causal effects. Let $\mathbf{W}$ be a adjustment set such that $\mathbf{W} \cup \mathbf{Z}$ satisfies the backdoor criterion for treatment $X$, outcome $Y$, and contexts $\mathbf{Z}$. The interventional distributions {$E[Y|do(X=x), \mathbf{Z}=\phi]$} and {$E[Y|do(X=x), \mathbf{Z}=\mathbf{z}]$} can be estimated, respectively, in terms of observational distribution with backdoor adjustment as:
\begin{equation}\label{eq:cateobs}
\begin{split}
    {E[Y|do(X=x), \mathbf{Z}=\phi] = \textstyle\sum_{\mathbf{w}}{E[Y|X=x, \mathbf{W}=\mathbf{w}]P(\mathbf{W}=\mathbf{w})}}, \text{and}\\
    {E[Y|do(X=x), \mathbf{Z}=\mathbf{z}] = \textstyle\sum_{\mathbf{w}}{E[Y|X=x, \mathbf{W}=\mathbf{w}, \mathbf{Z}=\mathbf{z}]P(\mathbf{W}=\mathbf{w}|\mathbf{Z}=\mathbf{z})}}.
\end{split}
\end{equation}
% We can estimate $X_{CATE}(\mathbf{Z}=\mathbf{z})$ by substituting two expectation terms in Equation \ref{eq:ate} with Equation \ref{eq:cateobs}. 
To satisfy backdoor criteria, $\mathbf{W}$ should not be descendants of $X$ and $\mathbf{W} \cup \mathbf{Z}$ should block all backdoor paths from $X$ to $Y.$  
These estimands can be substituted in Equation \ref{eq:ate} to estimate CATE.
% The identification conditions of CATE for various heterogeneous contexts, including latent heterogeneity, have been covered by ~\citet{pearl-smr15}. 
% Also, CSL should identify correct causes and potential effect modifiers.

\vspace{-1em}
\subsection{Characterization of Effect Modifiers with SCM}
Here, we characterize effect modifiers with SCM. In general, effect modification is induced due to the presence of direct interactions between context $Z\in\mathbf{Z}$ and treatment $X$ in the underlying function that defines outcome $Y$, e.g., $Y=f_Y(g(X)\times h(Z))$, where $f_Y \in \mathbf{f}$, $g$ and $h$ are feature mapping functions, and $\times$ denotes interaction. In the above example, the causal effect of $X$ on $Y$ can be estimated by taking a partial derivative of $Y$ with respect to $X$, i.e., $\frac{\partial Y}{\partial X}=f_Y'(.)h(Z)g'(X)$, where the causal effect of $X$ on $Y$ varies depending on $h(Z)$. The non-linearity of the function $f_Y$ can introduce heterogeneity, due to the term $f_Y'(.)$, even when there is no direct interaction. For example, if $Y = e^{3X + Z}$, then $\frac{\partial Y}{\partial X}=3e^{3X + Z}$, with $f_Y'(.)=e^{3X + Z}$ indicating variability in causal effects depending on $Z$. We refer the heterogeneity due to non-linearity of functions as indirect interaction.

\begin{figure*}[t]
    \centering
    \includegraphics[width=0.9\linewidth]{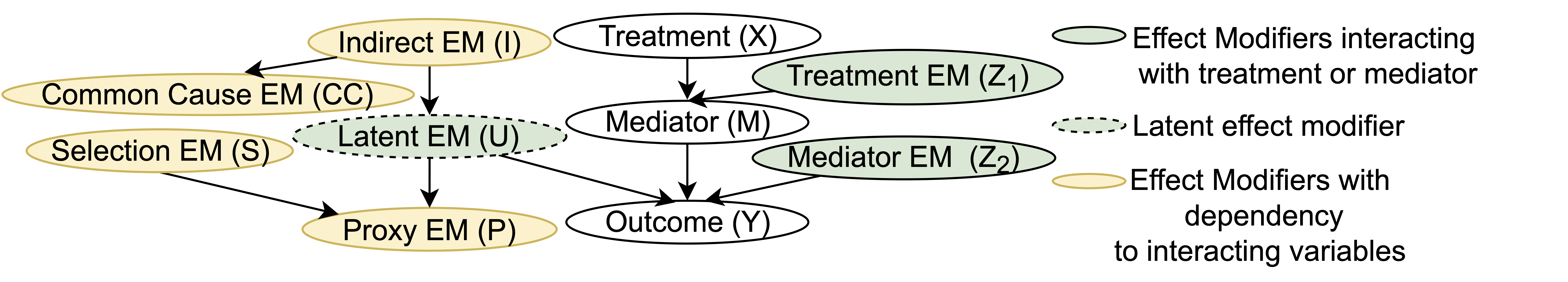}
    \caption{Characterization of effect modifiers (EM) with a structural causal model (SCM).}
    \label{fig:em}
    % \vspace{-1em}
\end{figure*}

Figure \ref{fig:em} summarizes our characterization of effect modifiers, adapted from \citet{vanderweele-epi07}, based on interactions and dependencies in $G$: (1) effect modifiers (EM) interacting with treatment or mediator, (2) latent EM, and (3) EM with dependency to interacting variables. A shared child node in $G$ may result in a direct or indirect interaction. While mediator EMs only share a child ($M$ or $Y$) with mediators and not treatment directly, treatment EMs share a child ($M$ or $Y$) with treatment.
These interacting variables could be latent, either known or unknown. As shown in Figure \ref{fig:em}, indirect, proxy, common cause, and selection effect modifiers are non-descendants of treatment with dependence on interacting variables via ancestor, descendant, confounding, and selection (or spouse) relationships, respectively. In Figure \ref{fig:em}, only the effect modifers $Z_1$, $Z_2$, and $I$ are causal effect modifiers, while the others ($P$, $S$, and $CC$) are non-causal. \rev{Purely data-driven methods for discovering treatment heterogeneity (e.g., ~\cite{athey-pnas16}) often optimize predictive separation and may select correlated surrogates as effect modifiers. Our model-driven effect-modifier discovery aims to mitigate spurious heterogeneity by prioritizing effect modifiers that interact causally with the treatment.}
% \vspace{-1em}
\subsection{\rev{Practical Considerations and Hypothesis}}
A natural approach to addressing Problem 1 is to learn a {$G(\mathbf{V}, \mathbf{E})$} to find causes, and Problem 2 is to identify potential effect modifiers, and then test the presence of heterogeneous effects. Causal structure learning (CSL)~\cite{spirtes-book00} concerns learning the adjacency and the orientation of the edges in {$G$} using observational data and, optionally, background knowledge. Some CSL algorithms like FCI~\citep{spirtes-book00} can even detect the presence of latent confounding. However, to infer a causal structure without accessing interventional distribution, CSL methods must make assumptions about the unknown underlying data-generating mechanism.
Causal sufficiency and faithfulness are the main assumptions made by CSL algorithms for estimating correct adjacencies and orientations of edges. These assumptions may be relaxed by some CSL algorithms. \textit{Causal sufficiency} assumes that there are no unmeasured common causes of variables in $\mathbf{V}$. \textit{Faithfulness} assumes all the conditional independencies (CI) observed from the data are entailed by the d-separation conditions~\cite{pearl-book88} of an underlying $G$.

In our problem, violating causal sufficiency can lead to false causes and non-identifiability of effect modification. For example, in Figure \ref{fig:em}, with unobserved $U$, the proxy effect modifier $P$ may be falsely discovered as a cause of $Y$ and a causal effect modifier. Even though our proposed framework is general enough to include CSL algorithms that allow for latent variables, for simplicity of exposition, we focus on those that assume causal sufficiency. Due to heterogeneity, the data distribution may exhibit more CI relations than those implied by $G$, thereby violating the faithfulness assumption. For example, a collider structure $X \rightarrow Y \leftarrow Z$ with binary $X=\theta(0.6), Z=\theta(0.5)$, and $Y=\theta(0.4+0.2X+0.24Z-0.4XZ)$ has an unfaithful distribution. Here, $E[\frac{\partial Y}{\partial X}]$ and $E[\frac{\partial Y}{\partial Z}]$ are both zero making both $X$ and $Z$ marginally independent of $Y$ although $X$ and $Z$ are causes of $Y$.
Although unfaithfulness due to counterbalancing parameters is rare~\cite{spirtes-book00}, a sufficient sample size and absence of selection bias are desired in practice for faithful distributions. CSL methods may rely on CI tests, goodness-of-fit scores, and cause-effect asymmetries, making additional parametric assumptions about the nature of the functions in SCM~\citep{glymour-fg19}.
% Even though our proposed framework is general enough that it can include CSL algorithms that allow for latent variables, for simplicity of exposition, we focus on ones that assume causal sufficiency.
% \newchange{CSL methods rely on CI tests, goodness-of-fit scores, and cause-effect asymmetries~\cite{glymour-fg19} making additional parameteric assumptions about the type of data distribution, such as continuous, discrete, or mixed, and the nature of the functional causal model.
% % , such as linear vs non-linear, or the type and distribution of exogenous noise.
% Moreover, most algorithms assume the data generation mechanism is homogeneous. However, each independent mechanism $P(V_i|pa(V_i, G))$ may have a distinct data generating process. CSL algorithms must be robust to heterogeneous data generation processes that use both linear and nonlinear functions in order to reliably discover causes and effect modifiers.
% } 

We focus on discovering causes and causal effect modifiers of the outcome. The data for our use cases is largely binary, indicating the presence or absence of a disease or an event. Without loss of generality, we focus on non-linear mechanisms for generating categorical data.
% , with inherent heterogeneity induced by non-linearity due to some underlying thresholding mechanism.
% We concentrate on two queries: outcome's direct parents $pa(Y,G)$ and ancestors $anc(Y,G)$.
One important difficulty in CSL is the uncertainty about whether a model's assumptions are violated when they are untestable with the data. Causal sufficiency, faithfulness, acyclicity, and absence of selection bias are mostly untestable. For our use case with largely binary variables, parametric assumptions like additive noise models are too restrictive or unsuitable~\cite{peters-pami11}. CSL algorithms may be sensitive to the number of nodes, edge sparsity, the underlying network topology, and the strength of causal effects. We hypothesize that an ensemble of CSL algorithms can mitigate this issue by discovering causes that persist across different assumptions. Here, we test whether multiple CSL algorithms in an ensemble agree on actual causes and make dissimilar mistakes for false causes, and test the utility of an ensemble for ranking causes based on confidence scores. 
While careful model selection is desirable, we lack a clear metric to identify a correct model from others under untestable assumptions and the absence of ground truth. We rely on empirically evaluating the CSL ensemble using simulated data. Theoretical support for the CSL ensemble approach is not trivial because the potential causes are not independent of one another.
% Theoretical support for the CSL ensemble approach is not trivial due to the non-iid potential causes. As a result, we rely on empirically evaluating the CSL ensemble using simulated data.

% For simplicity of exposition, we defer handling latent confounding to Appendix and focus on CSL algorithms under causal sufficiency assumption.
% Since different structure learning algorithms make diverse untestable assumptions about the data, we hypothesize an ensemble can discover which causes persist across different assumptions. In this work, we analyze the robustness of CSL algorithms and their ensemble for jointly discovering causes and effect modifiers from realistic data. We concentrate on the discovery of effect modifiers due to direct or mediated interaction with treatment. Even though our proposed framework is general enough that it can include CSL algorithms that allow for latent variables, for simplicity of exposition, we focus on ones that assume causal sufficiency.

%% file: 5-methodology.tex
% \vspace{-1em}
\section{Heterogeneous Causal Discovery Framework}\label{sec:methodology}
% \begin{figure*}[t]
% \vspace{-1em}
%     \centering
%     \includegraphics[width=0.9\linewidth]{images/csl.png}
%     \caption{Heterogeneous causal discovery framework.}
%     \label{fig:csl}
%     % \vspace{-1em}
% \end{figure*}

\begin{figure*}[t]
\vspace{-1em}
    \centering
    \includegraphics[width=0.9\linewidth]{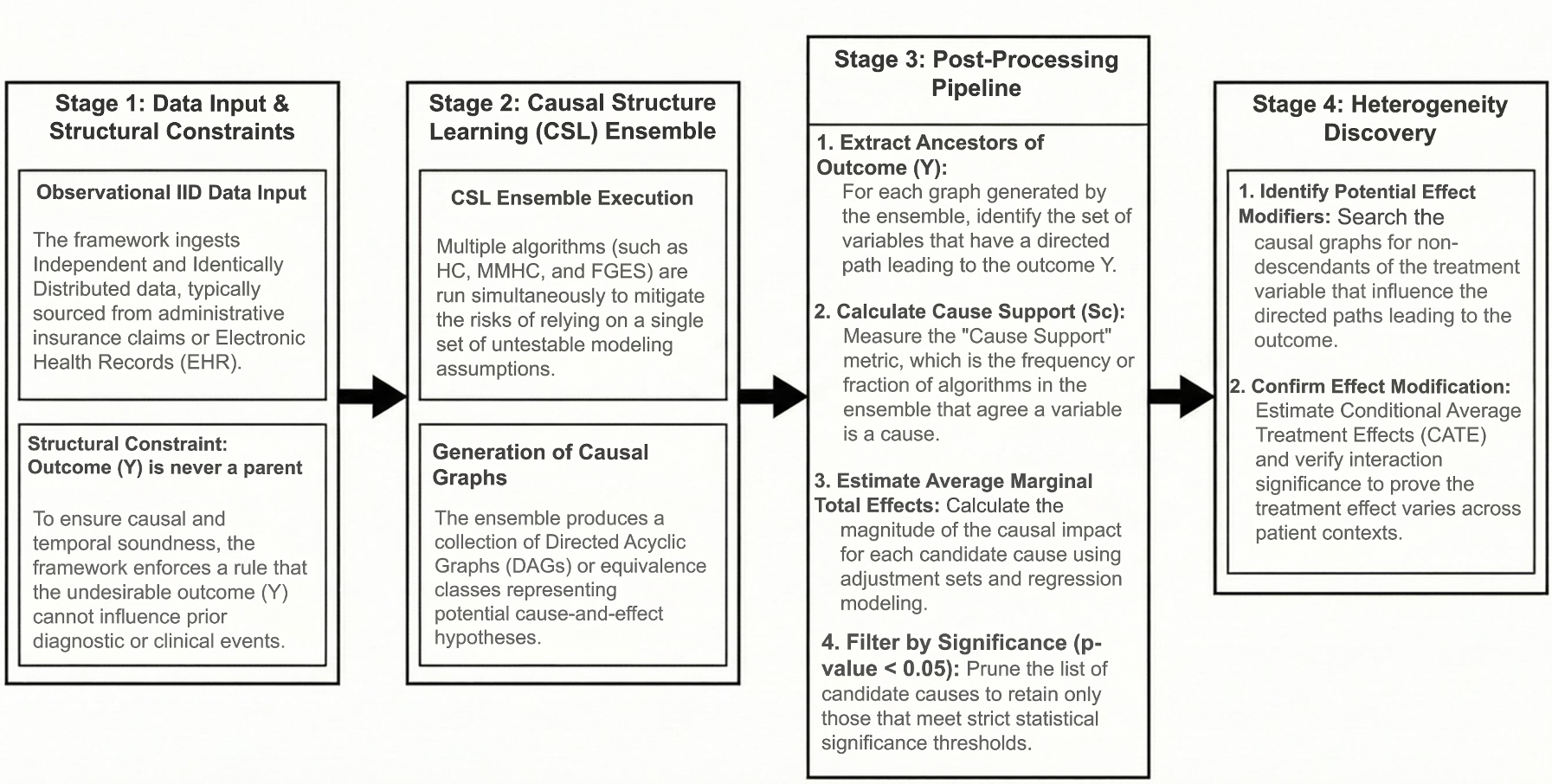}
    \caption{Workflow of heterogeneous causal discovery framework.}
    \label{fig:csl}
    % \vspace{-1em}
\end{figure*}

Figure \ref{fig:csl} presents a high-level overview of our proposed methodology for discovering heterogeneous causes, i.e., causes and causal effect modifiers, that are responsible for triggering or preventing the outcome. The framework takes independent and identically distributed (IID) observational data as an input. Then, an ensemble of CSL algorithms is employed to learn causal graphical models, which are used to discover prominent causes that trigger or prevent the outcome.
Finally, the module for heterogeneity analysis takes the discovered causes, observational data, and causal models to discover the effect modifiers of the outcome together with heterogeneous effects estimates. 
The data consists $N$ records \rev{each with M-dimensional} observed variables, \rev{i.e., $\mathbf{X}\in \mathbb{R}^{N\times M}$, and binary} outcome $Y \in \mathbb{B}^N$.
\subsection{Causal Structure Learning Ensemble}\label{sec:discoveringCauses}
Problem 1 of discovering causes, described in Section \ref{sec:problemsetup}, translates to learning {$G(\mathbf{V}, \mathbf{E})$} with vertices {$\mathbf{V}= \mathbf{X} \cup \{Y\}$}, and then answering {$anc(Y, G)$} query. This module takes observational data and \textit{structural constraints} as inputs. The CSL ensemble module outputs a list of causes and \textit{cause support} metric as a measure of confidence for each cause.

\textbf{Structural constraints.} 
Structural constraints are optionally used by CSL algorithms to encode background knowledge. 
We assume that the variables $\mathbf{X}$ are measured prior to the outcome $Y$. This is a reasonable assumption in real-world use cases. For instance, the diagnosis codes encountered prior to repeat ED visits are considered as potential causes. This also ensures the soundness of the CATE estimand, as discussed in Section \ref{sec:problemsetup}. Therefore, we include a structural constraint that the outcome $Y$ is never a parent, i.e., {$\forall {V_j|E(V_i\rightarrow V_j)} \in \mathbf{E}, Y \not\in pa(V_j, G)$}.
% \fixme{TODO: Move to experiments and appendix...}

\textbf{Cause support.}
The CSL ensemble takes the data and constraints as inputs to multiple CSL algorithms with possibly different causal assumptions or methodologies, and outputs DAG or equivalence classes (see Appendix A) of DAGs. The constraints aid in the orientation of edges adjacent to Y as well as the orientation of undirected edges in the equivalence class. To obtain causes of the outcome $Y$, we iterate through all $K$ output causal graphs and find tuples of a variable and a graph such that the variable is an ancestor of the outcome in the graph i.e.
{$\mathbf{R}:=\{(X_j, G_k)~|~\exists{G_k \in \{G_1,...,G_K\}}, X_j \in anc(Y, G_k) \wedge X_j \in \mathbf{X}$\}}.
The cause support for a cause $X_j$ shows the fraction of final $K$ graphs, one per CSL algorithm, that contain $X_j$ as a cause of outcome $Y$, and is defined as { $S_c(X_j)=\frac{1}{K}\sum_k{ \mathbb{1}[(X_j,G_k)\in \mathbf{R}]}$}, where $\mathbb{1}[.]$ is an indicator variable.

% \subsection{Heterogeneous and Total Effect Estimation}
\vspace{-1em}
\subsection{Discovering heterogeneity}\label{sec:discoverHte}
% \fixme{Handle interaction between backdoor adjustment set and conditional variable using linear model with interactions}
This module takes the observational data, $K$ output graphs, and the set $\mathbf{R}$ consisting of the tuples of a cause with a pointer to the corresponding graph as inputs.

\textbf{Total effect estimation.} We estimate the average total causal effect (ATE) {${X_j}_{CATE}(Z=\phi)$} for each pair of {$(X_j, G_k) \in \mathbf{R}$} using backdoor adjustment set {$\mathbf{W}=pa(X_j,G_k)$}, if identifiable. Nonparametrically, the ATE is calculated by substituting Equation \ref{eq:cateobs} into Equation \ref{eq:ate} and calculating the expectations and probability distributions. This approach maybe computationally expensive and, in practice, ATE can be estimated by regressing $Y$ on treatment and adjustment set $\{X_j\} \cup \mathbf{W}$ with any suitable machine learning model. We use a linear model to obtain ATE and its statistical significance utilizing the regression coefficient and p-value of $X_j$.

% The interventional distribution {\small $E[Y|do(X=x)]$} can be estimated in terms of observational distribution with backdoor adjustment as {\small $E[Y|do(X=x)] = \sum_w{E[Y|X=x, \mathbf{W}=\mathbf{w}]P(\mathbf{W}=\mathbf{w})}$}.
% The average total causal effect is obtained by nonparametric estimation or by regressing {$Y$} on {$\{X^j\} \cup \mathbf{W}$} with a linear model. The regression coefficient of {$X^j$} gives the average total effect.\\
% The post-processing module summarizes the causes and the estimated multi-set of causal effects consistent across output causal models. The causes are first ranked according to support of causal models i.e. {\small $\sum_k{\mathbbm{1}_\mathbf{R}(X^j,G_k)}$} and again ranked by magnitude of maximum absolute causal effect in a multi-set, as illustrated in Figure \ref{fig:results_ate}.
%The estimation with the above procedure does not capture the heterogeneity of causal effects in different sub-populations. 
% Next, we present our methodology to identify the sources of heterogeneity for a cause {\small $X^j$}, effect {\small $Y$}, and causal models {\small $\{G_1,...,G_K\}$}.
% \begin{figure}
%     \centering
%     \includegraphics[width=0.5\linewidth]{images/effect_modifier_demo.pdf}
%     \caption{Demonstration of potential effect modifiers. C and D are potential effect modifiers for treatment X and outcome Y.}
%     \label{fig:em_demo}
% \end{figure}
\textbf{Identifying effect modifiers.} 
% If the causal effect of a treatment variable {\small $X^j$} on an outcome {\small $Y$} is heterogeneous for different assignments of a variable or a set of variables {\small $\mathbf{Z}$}, then {\small $\mathbf{Z}$} is said to be an \textit{effect modifier} (or \textit{source of heterogeneity}) for the treatment {\small $X^j$} and outcome {\small $Y$}.
% In this work, we are interested in identifying the effect modifiers that are causes of the outcome.
The module for analysis of heterogeneity, shown in Figure \ref{fig:csl}, involves short-listing potential effect modifiers (EM), for a given treatment {$X_j$} and outcome {$Y$}, together with estimation of the heterogeneous and average total treatment effects across causal models {$\{G_1,...,G_K\}$}.
% \textbf{Short-listing potential effect modifiers}. %\citet{vanderweele-epi07} introduce a theorem to identify potential effect modifiers for a given treatment and outcome in a DAG. 
% The potential effect modifiers for a given treatment and outcome in a DAG include the non-descendants of the treatment node and either (1) the parents of the outcome or (2) the parents of mediator nodes in the directed path from treatment to outcome~\cite{vanderweele-epi07}.
%In Figure \ref{fig:em_demo}, the nodes $C$ and $D$ are potential effect modifiers for treatment $X$ and outcome $Y$ as these nodes are non-descendants of $X$ and parents of outcome and mediator respectively.
Let {$G^{anc}_k(\mathbf{V}^{anc},\mathbf{E}^{anc})~|~V^{anc} \in \{Y\}~\cup~anc(Y, G_k)$} denote a sub-graph of {$G_k$} with {$Y$}, {$anc(Y,G_k)$}, and edges between them retained. Given a graph {$G_k$}, treatment {$X_j$}, and outcome {$Y$}, the potential treatment EM and mediator EM, as described in Section \ref{sec:assumptions}, are tuple of node and graph pair {$\mathbf{Z}:=(Z_i, G_k)~|~Z_i \notin des(X_j, G_k^{anc}) \land \exists{V_d \in des(X_j, G_k^{anc})}, Z_i \in pa(V_d, G_k^{anc})$}. 
%The potential effect modifiers are identified for all causal models $G_k \in \{G_1,...,G_K\}$.

\textbf{Estimating heterogeneous treatment effects (HTE).} We check the presence of actual heterogeneity for treatment {$X_j$}, outcome {$Y$}, and potential effect modifier {$(Z_i, G_k) \in \mathbf{Z}$} following Equation \ref{eq:hte}. 
The estimation of CATE uses backdoor adjustment set {$\mathbf{W}$} conditioned on {$Z_i=z$}.
% The interventional distribution {\small $E[Y|do(X^j=x), Z_i=z]$} in Equation \ref{eq:ate} is estimated in terms of observational distribution as {\small $E[Y|do(X^j=x), Z_i=z] = \sum_w E[Y|X=x, W=w, Z_i=z]P(W=w|Z_i=z)$}.
Similar to ATE, the HTE is estimated nonparametrically or by regressing {$Y$} on {$\{X_j,Z_i,Z_i \times X_j\} \cup \mathbf{W} \cup \{\mathbf{W} \times Z_i\}$}, where {$Z_i \times X_j$} is an interaction term between treatment and conditional variable and $\mathbf{W} \times Z_i$ captures interactions between adjustment set and conditional variable. %\footnote{We handle the edge case $Z\subset\mathbf{W}$}. 
Using a linear regression model with interaction terms, we can detect the presence of a heterogeneous effect with a statistically significant non-zero coefficient for the interaction term. Here, we demonstrate identifying heterogeneity using a linear model with pairwise interaction terms. However, the variables $\{X_j, Y, \mathbf{W}, \mathbf{Z}\}$ can be used with any HTE estimator~\citep{kunzel-pnas19,athey-pnas16} to discover more complex subpopulations.
% We also handle the edge case $Z\subset\mathbf{W}$ for both estimation methods.

The final output of the framework includes the following: (1) for an observational dataset and an outcome, a list of causes with causal support and a multi-set of their average total causal effects, and (2) for a given cause and outcome, a list of effect modifiers with causal support and a multi-set of heterogeneous treatment effects.

%% file: 6-experiments.tex
\section{Experiments and Results}\label{sec:experiments}
We conduct experiments to evaluate the proposed framework for discovering causes and sources of heterogeneity for an outcome \rev{in multiple ways.} First, using two real-world data, we demonstrate that the top causes triggering or preventing the outcome along with effect modifiers, discovered by our approach, are consistent with the published research literature and common knowledge. Second, we evaluate the significance of the ensemble approach using synthetic and semi-synthetic data. \rev{Then, we demonstrate portability our framework to modern EHR datasets and test whether the insights found for vulnerable, high-risk population generalizes to general population in a specific healthcare settings. Finally, we report the findings of validation by four clinicians for a short and structured survey designed to evaluate important causal-relationships identified by a reliable baseline and the majority ensemble.}
% \vspace{-1em}

\subsection{Real-World Application: Repeat ED Visits for Diabetic Population}\label{sec:expreal}
Here, we describe the real-world dataset and experimental setup for discovering heterogeneous causes of repeat ED visits.
We present results summarizing the discovered causes and effect modifiers for several potential interventions. Because of the task novelty and \rev{limited} ground truth, we \rev{first} validate the discovered heterogeneous causes through published medical research. 
We use statistically de-identified medical claims data provided by a leading US-based health insurance company for repeat ED visit analysis. One advantage of insurance claims over electronic health records (EHR) is that they contain information from multiple healthcare providers.
The dataset consists of $461,754$ claims from the year 2015 to 2018 for $11,893$ diabetic patients with at least one ED visit and one or more claims after the first ED visit. \rev{We define outcome $Y$ as} whether or not a patient has a repeat ED visit within $12$ months of the first visit. \rev{We choose recurrence over a 12-month horizon because our objective is to capture repeat ED utilization relevant to chronic disease burden, ongoing disease management and care access, and outpatient utilization patterns after the incident event. By contrast, prior work typically defines frequent flyers using annual visit counts (e.g., $\ge4$ ED visits/year)~\cite{lacalle-aem10,soril-plos15} to characterize established high utilizers and their links to factors like chronic disease burden, insurance coverage, and outpatient care access.  Much shorter revisit windows (e.g., 48–72 hours) are typically used as ED quality or safety metrics and emphasize immediate post-discharge issues rather than longer-run disease control~\cite{liu-ijcm25,ratti-shti24}.} Empirically, we observe from the dataset that nearly $80\%$ of patients with multiple ED visits have a second ED visit within $12$ months of the first one. The dataset consists of $45\%$ patients with a repeat ED visit ($Y=1$) within $12$ months.

% %Each claim for a patient is annotated as either ``ED claim'' or ``Non-ED claim'' and the dataset consists of 10\% ED claims.
% We consider the ICD-10 diagnosis codes in the medical claims as potential causes of the repeat ED visits. %The time granularity of each claim is per day. 
% %The granularity of ICD-10 codes is 4 digits except for the Z79 (long-term drug therapy) family with 6 digits granularity.
% We consider $80\%$ of most frequent ICD-10 codes and a binary out-of-vocabulary (OOV) indicator that captures the remaining less frequent ICD-10 codes among all diabetic populations. This allows us to limit dimensionality of features and reduce the risk of spurious associations due to rare codes with low prevalence among patients. To merge similar concepts and further reduce dimensionality, the granularity of ICD-10 codes is set to 4 digits except for the Z79 (long-term drug therapy) family with 6 digits granularity.

We consider the ICD-10 diagnosis codes in the medical claims as potential causes of the repeat ED visits. \rev{We retain the top 80\% most frequent ICD‑10 codes among all diabetic populations and collapse the remaining long‑tail codes into a single binary out‑of‑vocabulary (OOV) indicator. This design to limit feature dimensionality is important to keep causal discovery computationally feasible and to avoid unstable associations driven by very rare codes. To further reduce dimensionality while preserving clinical meaning, we truncate ICD‑10 codes to 4 characters, except for the Z79 (long‑term drug therapy) family, which we retain at 6-character granularity to distinguish clinically important medication subtypes.} 
We flatten the claims of each patient using binary count aggregation\rev{, resulting in a 335-dimensional feature vector $\mathbf{X}$,} that indicates the presence or absence of a diagnosis code. We ensure that only the diagnosis codes encountered prior to measuring the outcome are aggregated. If $Y = 1$, the diagnosis codes before the second ED visit are aggregated, and if $Y = 0$, diagnosis codes up to $12$ months after the first ED visit are aggregated.

\textbf{Results.}
For the real-world experiments, we consider three CSL algorithms in the ensemble module: %Greedy Equivalence Search (GES)~\cite{chickering-jmlr02},
HC~\cite{scutari-jss10}, FGES~\cite{ramsey-ijdsa17}, and MMHC~\cite{tsamardinos-ml06}. The choice of CSL algorithms is motivated by the high dimensionality of the variables and by their parallelized implementation. \rev{Furthermore, the performance of these algorithms was reliable, even compared to modern methods, on extensive simulated data experiments (which we present in Section \ref{sec:simulation}).
%All algorithms make assumptions of causal sufficiency and causal faithfulness. However, 
% FGES's implementation in \texttt{py-causal}~\cite{pycausal-github19} package provides robustness to faithfulness assumption. For HC and MMHC, we use \texttt{bnlearn}~\cite{scutari-jss10} package. 
The output for MMHC and HC is a directed acyclic graph (DAG), and for FGES is a completed partially directed acyclic graph (CPDAG). We use the temporal order of diagnosis codes as an orientation support heuristic to orient a few undirected edges in the CPDAG.
% We should note that the order of disease progression and diagnosis may not always be the same because diagnoses are like pieces of a puzzle that come together to understand a patient’s needs.
The Appendix provides details on the orientation support heuristic, the algorithm's assumptions, and the hyperparameters.}

\begin{figure*}[t]
        \centering
        \subfigure[Top 25 ICD-10 codes with positive effect.] {%{0.49\linewidth}
    \includegraphics[width=0.46\linewidth,keepaspectratio]{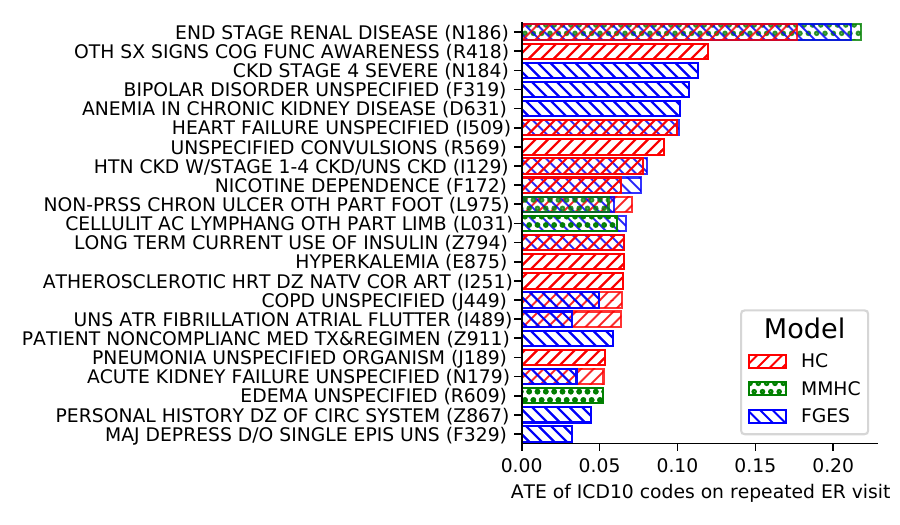}\label{fig:pos_ate}
    % \caption{Top 25 ICD10 codes with positive effect.}
    }
    % \hfill
    \qquad
    \subfigure[Top 25 ICD-10 codes with negative effect.]{%{0.49\linewidth}
        \centering
         \includegraphics[width=0.46\linewidth,keepaspectratio]{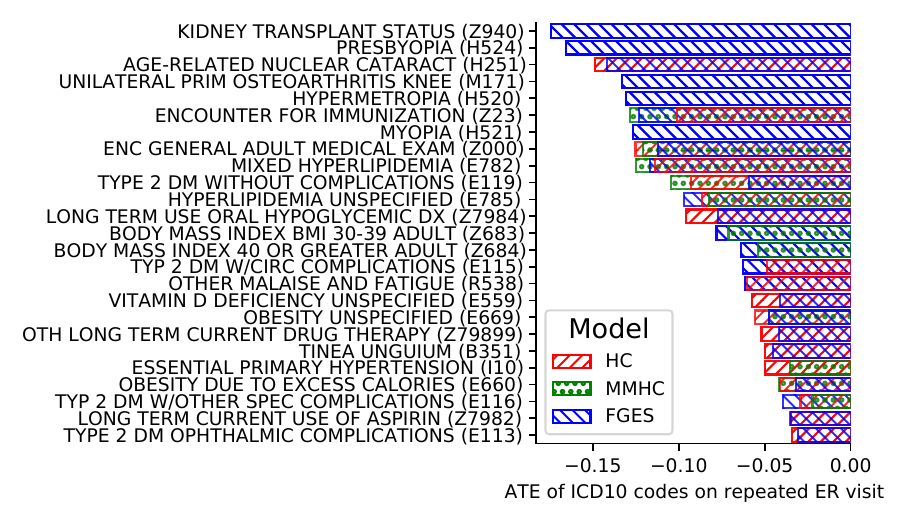} \label{fig:neg_ate}
        % \caption{Top 25 ICD10 codes with negative effect.}
}
    \caption{Multi-set average total causal effect of top diagnosis codes on the repeat ED visits.}
    \label{fig:results_ate}
    \vspace{-1.2em}
\end{figure*}

Figure \ref{fig:results_ate} depicts the multi-set of average total causal effects (ATE) for top ICD-10 diagnosis codes on the repeat ED visits. 
Figure \ref{fig:pos_ate} shows top diagnosis codes with positive ATE that are indicators of risk factors for increased likelihood of repeat ED visits. Similarly, Figure \ref{fig:neg_ate} shows top diagnosis codes that are likely related to preventive factors of repeat ED visits. Next, we summarize some interesting results produced by our approach and validate them using published research.
% \textbf{\emph{Kidney complications are major risk factors of repeat ED visits}}.  Other kidney complications are comorbidities for diabetic patients and can ultimately lead to kidney failure~\cite{weiner-jmcp07}.

\emph{Patients with kidney, circulatory, respiratory, and mental health complications are vulnerable to repeat ED visits}.
%We observe kidney complications (N186, D631, I129, N179, N184) have positive effect on the repeat ED visits for the diabetic population.
%Kidney failure (N186) is consistent across all causal models.
This result is consistent to the model of risk factors for frequent ED visits validated by the Emergency Medicine physicians~\citep{birmingham-ajem20}.
Kidney failure and dependence on dialysis are identified to be the causal factors for heavy ED utilization~\cite{komenda-plos18}.
% We notice complications related to circulatory systems (I509, I489, I251, Z867), respiratory systems (J449, J189), and brain or mental health conditions (G20, R418, F319, F329) have positive effects on the repeat ED visits consistent to the model of risk factors for frequent ED visits validated by the Emergency Medicine physicians~\cite[p. 84]{birmingham-ajem20}.

% \textbf{\emph{Patients with advanced diabetes conditions are in the risk of repeat ED visits.}}
% Limb complications (L975) and long-term use of insulin (Z794) are the proxies for an advanced diabetes condition~\cite{said-nature07}. Studies show mismanagement of blood glucose and insulin leads to serious conditions requiring ED visits~\cite{kitabchi-dc09,grote-fns16}.

% \textbf{\emph{Patients with acute medical emergency symptoms are more likely for recurrent ED visits.}}
% We observe generic medical emergency symptoms~\cite{acep-online21} like convulsions(R569), edema (R609), and high potassium in the blood (E875) are identified as causal factors for repeat ED visits for the diabetic patients.

\emph{Interventions on patient noncompliance, insulin management, and smoking dependence can be further investigated to reduce repeat ED visits in diabetic population.}
Long-term use of insulin and patient noncompliance are discovered as risk factors for repeat ED visits consistent with previous studies~\cite{kitabchi-dc09,grote-fns16}. % and commonsense knowledge. 
% Interventions targeted to these vulnerable populations should be investigated.
Nicotine dependence is likely a factor interacting with other causes because %and its heterogeneity should be investigated~\cite{katz-acm12}. 
smoking is found to be associated with increased blood glucose levels and risk of cardiovascular and kidney comorbidities for diabetic patients~\cite{tonstad-drcp09}.

\emph{Interventions encouraging regular checkups, immunization, and drug adherence can be investigated for reducing repeat ED visits.} Figure \ref{fig:neg_ate} shows the ICD-10 codes with negative causal effects which can be considered as possible interventions or preventive factors. While some codes, such as Myopia, are not suitable as interventions, others, such as ``Encounter of General Medical Exam (Z000)” and ``Encounter for Immunization (Z23)” are. These preventive measures are recommended to diabetic population to prevent complications~\cite{smith-cd00,cdc-online21}. The ICD-10 codes associated with long-term (current) use of drugs are likely proxies for drug adherence. 
Other diagnosis codes are likely proxies for medical service use or effect modifiers capturing heterogeneity of plausible preventive factors.

\begin{figure*}[t]
        \centering
        \subfigure[EM for General Medical Exam (Z000).]{%{0.49\linewidth}
    \includegraphics[width=0.46\linewidth,keepaspectratio]{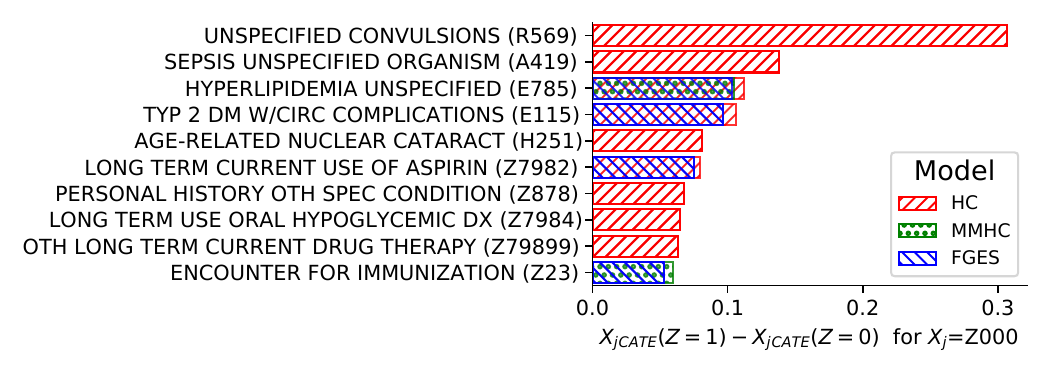}
    \label{fig:office_hte}
    }
    % \hfill
    \subfigure[EM for Immunization (Z23).]{%{0.49\linewidth}
        \centering
         \includegraphics[width=0.46\linewidth,keepaspectratio]{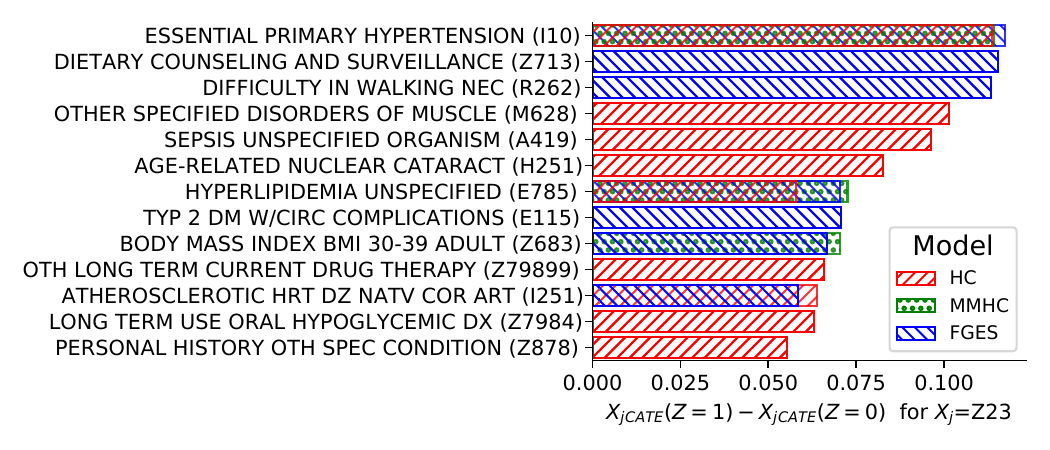}
        \label{fig:imm_hte}
    }
    \caption{Effect modifier (EM) of two preventive diagnosis codes on repeat ED visits.}
    \label{fig:results_hte}
    \vspace{-1.2em}
\end{figure*}

\emph{Potential interventions and heterogeneous causal effects}. 
% An intervention is not possible for all the potential factors triggering and preventing the repeat ED visits. 
% We analyze the sources of heterogeneity for the two ICD-10 codes, Z000 and Z23, related to preventive care that can likely be intervened. 
Figure \ref{fig:results_hte} depicts the ICD-10 diagnosis codes identified to produce heterogeneous causal effects for the two practical interventions, Z000 and Z23. The values in the x-axis are the differences of CATE of treatment {$X_j$} on repeat ED visits $Y$ for patients with and without a diagnosis {$Z$} i.e. {${X_j}_{CATE}(Z=1) - {X_j}_{CATE}(Z=0)$}. In Figure \ref{fig:office_hte}, we observe that intervention on the medical exam has less preventive effectiveness to patients with symptoms of convulsions (R569) compared to patients without convulsions. 
% Hyperlipidemia (E785) is a source of heterogeneity for both interventions consistent across all causal models.
Figure \ref{fig:results_hte} suggests interventions on both preventive measures for reducing repeat ED visits are less effective for diabetic patients with metabolic syndrome (primary hypertension (I10) or Hyperlipidemia (E785)) as well as for patients already utilizing health services such as drug use or dietary counseling. %Further investigation is necessary to ensure the feasibility of these preventive factors as potential targeted interventions for reducing repeat ED visits for diabetic patients.
% We defer the results of hospital readmission analysis to the Appendix B due to limited space.

\rev{We additionally run modern gradient-based causal discovery methods as baselines for the repeat ED visit analysis in the diabetic cohort, using NOTEARS (nonlinear)~\cite{zheng-aistats20}, GOLEM~\cite{ng-neurips}, and GraNDAG~\cite{lachapelle-iclr20} available in the gCastle package~\cite{zhang-arxiv21}. Across multiple hyperparameter settings (e.g., number of iterations, weight thresholds, and regularization parameters), NOTEARS (nonlinear) produced an empty graph (no edges). GOLEM inferred 117 directed relationships overall, but the repeat ED visit outcome was returned as an isolated node with no identified ancestors. GraNDAG inferred a total of 368 directed relationships, but the repeated ED visit outcome was incorrectly identified as the direct cause of two conditions: ``Diverticular Disease of Large Intestine without Perforation or Abscess (K573)" and ``Presence of a Prosthetic Heart Valve (Z952)". These behaviors suggest limited stability of these baselines in real-world setting~\cite{kaiser-springer22,olko-icml25}, and we use them only as supplementary comparisons.}

\subsection{Real-World Application: Hospital Readmissions for ICU Population}
For the hospital readmissions analysis, we use the freely available MIMIC-IV dataset~\cite{johnson-mimiciv21} that is sourced from two in-hospital database systems,  a custom hospital-wide EHR and an ICU-specific clinical information system.
% %,  for patients admitted to a tertiary academic medical center in Boston, MA, USA.
The dataset consists of de-identified data with diagnosis codes, admission records, and patient information from 2008 to 2019. For our study, we extract patient characteristics, hospitalization metadata, and the diagnosis codes with the ICD-9 version that gives $335,378$ hospitalizations records for $173,034$ patients. We set $30$ days as the time horizon for readmission analysis, consistent with prior studies~\cite{bailey-hcup19,balasubramanian-jamanet25}. The dataset consists of $17\%$ patients with at least one readmission within 30 days of discharge ($Y=1$).
We select potential causes as 70\% most frequent diagnosis codes and an OOV indicator for the remaining less frequent codes, using the same rationale as the repeat ED visit analysis. These codes are aggregated for each patient following a process similar to repeat ED visit analysis. For hospital readmission analysis, the data $\mathbf{X}$ includes patient characteristics such as age group, admission metadata such as previous discharge location, and diagnosis codes. \begin{figure*}[!t]
        \centering
        \subfigure[Top 25 ICD-9 codes with positive effect.]{
    \includegraphics[width=0.45\linewidth,keepaspectratio]{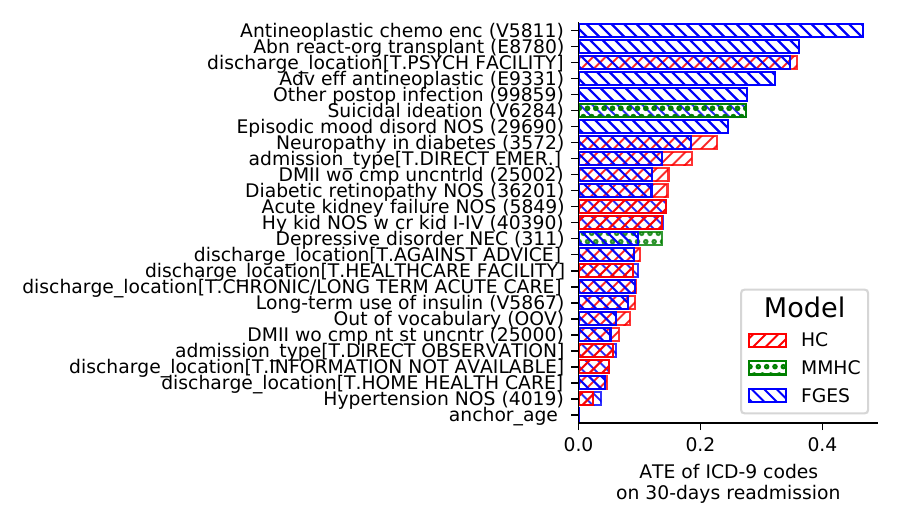}
    \label{fig:pos_atereadmit}
    }
    \qquad
    \subfigure[Top 25 ICD-9 codes with negative effect.]{
         \includegraphics[width=0.45\linewidth,keepaspectratio]{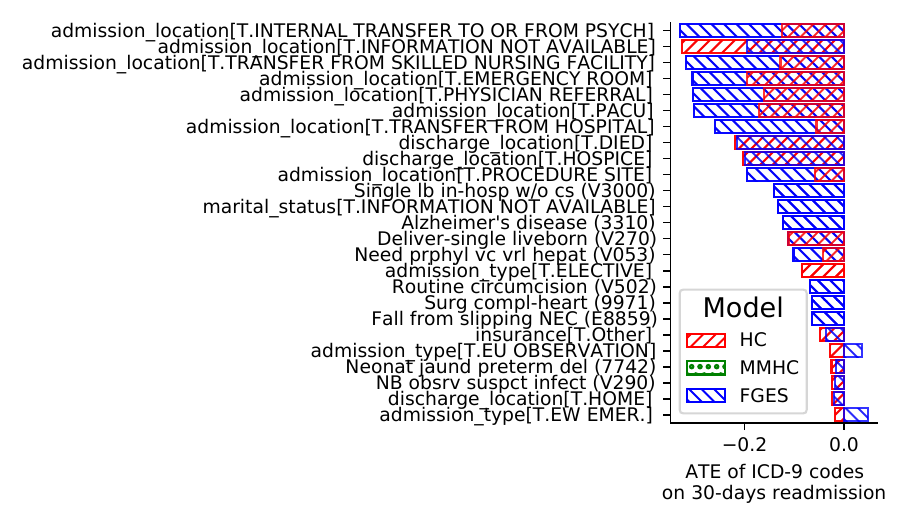}
        \label{fig:neg_atereadmit}
    }
    \caption{Multi-set total causal effect for top diagnosis codes on 30-day readmissions identified by three CSL algorithms.}
    \label{fig:results_atereadmit}
    \vspace{-1em}
\end{figure*}

\textbf{Results}. Figure \ref{fig:results_atereadmit} depicts the multi-set of total causal effects for the top ICD-9 diagnosis codes as well as other metadata on 30-day readmission. 
Figure \ref{fig:pos_atereadmit} shows top diagnosis codes and other contexts with positive total causal effects indicative of risk factors. Figure \ref{fig:neg_atereadmit} shows the top diagnosis codes and contexts that are potential preventive factors of readmissions. Here, we outline some interesting results produced by our approach consistent with published research~\cite{kripalani-arm14,benjenk-jhmhp18}.

\emph{Surgical complications, adverse effects of drugs, and advanced chronic diseases are risk factors for readmissions.} Figure \ref{fig:pos_atereadmit} shows diagnosis codes related to surgical complications (E8780, 99859) and adverse effects of drugs (E9331) has significant ATE. Similarly, the codes linked to chronic conditions like diabetes (3572, 25002, 36201, V5867, 25000), kidney failure (5849, 40390), hypertension (4019, 40390), and cancer (V5811, E9331) are positive, suggesting these subpopulations are vulnerable to readmissions.

\emph{Interventions targeting patients with mental health disorders should be investigated.} In Figure \ref{fig:pos_atereadmit}, we notice the diagnosis codes related to suicidal ideation (V6284), episodic mood disorder (29690), and depressive disorder (311) have positive ATE.

\emph{Heterogeneity of readmissions based on admission and discharge locations should be further investigated for designing interventions.} Figure \ref{fig:pos_atereadmit} and \ref{fig:neg_atereadmit} show various admit and discharge locations having diverse effects on readmissions. Patients admitted to the hospital directly (DIRECT  EMER. and DIRECT OBSERVATION) are more likely to end up being readmitted within 30 days than patients admitted to other locations. Interestingly, patients discharged to psychiatric, healthcare, and chronic/long-term acute care facilities are more prone to readmissions. Patients discharged against advice have a higher readmission rate, while those discharged to hospice are discovered to be less likely for 30-days readmission.
Additional investigation by stakeholders is needed to understand these behaviors.

\emph{Early diagnosis of risk factors can prevent readmissions.} Figure \ref{fig:neg_atereadmit} shows the quality of care such as the diagnosis of surgical cardiac complications (9971), observation for suspected infections (V290), and diagnosis of need for preventive vaccination (V053) reduces the likelihood of readmission.

\rev{\subsection{Real-World Application: Portability and Findings on the EHRSHOT Dataset}}
\rev{The previous two subsections establish the framework’s utility for generating actionable hypotheses within distinct high-risk subpopulations. Here, we apply the proposed framework to EHRSHOT~\cite{wornow-neurips23}, a general-population EHR cohort, to demonstrate portability across datasets with different risk profiles and across modern EHR representations standardized to the Observational Medical Outcomes Partnership (OMOP) Common Data Model (CDM)~\cite{ohdsi2019book}. EHRSHOT is a longitudinal, general-population EHR cohort containing de-identified structured records for 6,739 Stanford Medicine patients, comprising 41.6 million clinical events across 921,499 encounters/visits. The dataset excludes patients $<19$ or $>88$ years of age and retains only those with $\ge10$ recorded clinical events to ensure sufficient longitudinal history. We note that EHRSHOT was released as a few-shot prediction benchmark with a pretrained foundation model, not for causal discovery. We use it here only to demonstrate OMOP‑CDM portability and to contrast causal hypotheses about drivers of healthcare utilization across populations with different risk profiles.}

\rev{\textbf{Cohort and outcomes.} We identified 3,020 patients in the EHRSHOT dataset with at least one emergency department (ED) visit. Among these, 1,366 (45\%) experience at least one repeat ED visit within 12 months of the first ED visit (our repeat ED visit endpoint). For readmissions, we identify $3,718$ unique patients with at least one hospital admission, of which $269$ (7\%) have a 30-day readmission following the first discharge (our readmission endpoint). The cohort and outcome construction mirror our primary analysis for comparing dynamics across the high-risk and general populations.}

\rev{\textbf{Preprocessing and feature construction.} To keep the feature space tractable while preserving clinical meaning, we select the top 500 most frequent OMOP condition concepts in the overall EHRSHOT population. All remaining condition concepts are mapped to the closest available ancestor within this top-500 set using the OMOP concept hierarchy (ancestor/descendant relationships). All concepts that cannot be mapped to existing vocabulary are assigned to an out-of-vocabulary (OOV) bucket. Patient-level features are binary indicators of whether each condition concept occurred prior to the terminal endpoint (i.e., before the repeat event if it occurs, or before the end of the specified time horizon otherwise). We additionally include basic demographics (age, sex, race, and ethnicity) and append utilization history, capturing prior office visits and immunizations within the same pre-outcome window.}

\rev{\textbf{Results for repeat ED analysis}. In EHRSHOT, applying the same pipeline with three CSL algorithms yielded no candidate drivers or preventive factors for 12‑month repeat ED visits, i.e., the repeat‑ED outcome remained an isolated node without ancestors in all learned graphs. In substantive terms, this pattern is consistent with a setting in which repeat ED visits in a general, mixed-risk population are influenced by heterogeneous and potentially idiosyncratic factors (e.g., episodic acute events or unmeasured social/behavioral triggers) rather than a small set of dominant, consistently measurable clinical antecedents. At the same time, this null result should be interpreted cautiously, as the EHRSHOT cohort is relatively small for high-dimensional causal discovery, limiting power to detect weak effects. As a simple screen for measurable signals, we computed absolute Pearson correlations between the outcome and all candidate features. The distribution of correlations was near zero overall (n = 512; mean = 0.0046, SD = 0.0256; min = -0.1031, 25\% = -0.0113, median = 0.0047, 75\% = 0.0191, max = 0.1250), consistent with weak linear associations at the marginal level.}

\rev{\textbf{Results for readmission analysis}. In the readmission analysis, the FGES algorithm identified complications due to cancer (Malignant neoplastic disease and Anemia in neoplastic disease) and transplant (Disorder affecting transplanted structure) as prominent risk factors for 30-day readmission. Pancytopenia (due to chemotherapy), a serious blood condition characterized by a deficiency in all three cellular components of the blood, was identified as a common cause of anemia in neoplastic disease and the disorder affecting transplanted structures. The HC algorithm partially corroborated this by also selecting malignant neoplastic disease and out-of-vocabulary indicator as ancestors of 30-day readmission, whereas MMHC returned no ancestors. Notably, this thematic signal is consistent with the MIMIC-IV ICU cohort, where FGES also surfaced closely related factors, Antineoplastic chemotherapy encounter (V58.11) and Abnormal reaction organ transplant (E878.0), among the top readmission-associated codes. The detailed prior admission and discharge metadata were not readily available and were not included in the analysis of the EHRSHOT data.
}

\subsection{Validation with Synthetic and Semi-Synthetic Data Experiments}\label{sec:simulation}
We perform synthetic and semi-synthetic experiments to evaluate the benefits of the CSL ensemble over using a single algorithm.
% Additionally, we test the effectiveness of the ensemble approach on enriching the potential causal hypotheses. 
% The advantage of using simulated data is that it contains ground truth.
%A semi-synthetic experimental setup is necessary for the evaluation of causal discovery methods because there is a lack of ground truth in real-world datasets. 
% However, using purely synthetic data for evaluating causal models has limitations like high “researcher degrees-of-freedom” and lack of standardization~\cite{gentzel-nips19}.
%Moreover, synthetic data generation representing the phenomena of repeated undesirable outcomes is non-trivial.
% A semi-synthetic data generation strategy, that utilizes features of real-world data, helps in mitigating some of these limitations.
% Therefore, we employ a semi-synthetic data generation method to evaluate the benefits of our causal structural learning ensemble over using a single CSL algorithm. %The goal of our experiments is to assess the benefits of the ensemble approach compared to a single model selection strategy, especially when assumptions of the CSL algorithms are untestable. 
The experiments involve four steps: (1) generating ground-truth causal models, (2) generating (semi-)synthetic data from the models, (3) running experiments with the ensemble framework, and (4) evaluating the results.
% A detailed description of simulated DAGs and data generation processes is included in Appendix B.
% Here, we summarize DAG and data generation, and then present our main takeaways from the evaluation of simulated experiments with an ensemble of MMHC, HC, FGES, PC~\cite{spirtes-book00}, and LINGAM~\cite{shimizu-jmlr06} algorithms.

\textbf{Synthetic setup.} We want to test how sensitive CSL algorithms are to structural properties like the number of nodes, the sparsity of edges, and the network topology in discovering the causes of a sink outcome node. Similar to~\citet{zheng-aistats20}, we generate ground-truth directed acyclic graphs (DAGs) according to Erd{\H{o}}s R{\'e}nyi ($ED$)~\citep{erdos-pmihas-60} and Barab{\'a}si Albert ($ba$)~\citep{albert-rmp02} random graph models by varying the number of nodes and connectivity parameters.
The undirected degree distributions of $ED$ and $ba$ graphs are binomial and power law due to random and preferential attachment edge adjacency, respectively.
Both graphs are oriented with random topological order, ensuring the outcome $Y$ is never a parent. We defer the details to the Appendix.

We consider three modes of synthetic data generation: $logistic$, $logistic+interaction$, and $Bernoulli~linear$. The effect modification between variables due to non-linearity alone and due to both interaction and non-linearity is respectively captured by the $logistic$ and $logistic+interaction$ data generations. There is no effect modification in the $Bernoulli~linear$ data generation, and the size of the causal effects may be small. Appendix presents the detailed functional forms.

\textbf{Semi-synthetic setup.} In addition to randomly generated graphs, we use real-world data to generate ground truth causal graphs and learn their SCM parameters. We use a base algorithm to learn a causal graph and its parameters, which we consider as the ground truth. We present an analysis using MMHC as the base algorithm, but we observe similar trends with other base algorithms. Top $N \in \{10,20,30,40,50,75\}$ most frequent diagnosis codes, along with the outcome, are selected from the aggregated data, described in Section \ref{sec:expreal}, for generating ground-truth graphs of different sizes. The top N codes include an out-of-vocabulary (OOV) variable indicating the presence of other less frequent diagnosis codes. This data is used to learn a ground-truth causal model $G_T(\mathbf{V_T}, \mathbf{E_T})$. If any edges in the causal model are unoriented, we use orientation support, as defined in the Appendix, to orient them. From the ground-truth causal graph, we estimate the SCM parameters as conditional probability tables (CPT) $p(V_i|pa(V_i, G_T))$. Using the CPT and the topological ordering of the ground truth causal DAG, we sample IID semi-synthetic data.

\textbf{Results.} Next, we present the main takeaways from the evaluation of the experiments with an ensemble of MMHC, HC, FGES, PC, and LINGAM~\cite{shimizu-jmlr06} algorithms.

%\textbf{Experiments}.
% \textbf{Evaluation metrics}.
% We run an ensemble of five CSL algorithms -- MMHC, HC, FGES, PC, and LINGAM -- on the semi-synthetic data. 
% To assess the performance of the ensemble approach in refining potential causal factors, we estimate the average causal support for the true causes ($TP(S_c)$) and the false causes ($FP(S_c)$) defined as
% % \begin{equation}
%     $TP(S_c) = \frac{\sum_{X^j \in anc(Y, G_T)}S_c(X^j)}{n(anc(Y, G_T))}, \text{and}$
%     $FP(S_c) = \frac{\sum_{X^j \notin anc(Y, G_T)}S_c(X^j)}{n(G_T) - n(anc(Y, G_T))}$,
% % \end{equation}
% where $Y$ is the outcome and $S_c(X^j)$ is the cause support for event $X^j$. The $TP(S_c)$ and $FP(S_c)$ measures indicate the agreement among multiple CSL algorithms for true causes and false causes, respectively. In addition to the structural cause support, we investigate parametric significance support by checking the fraction of algorithms with statistically significant (p-value < $0.05$) total effects for the identified causes. The significance support gives a measure of confidence for discovered causes. Moreover, we compare individual algorithms and the ensemble in terms of true-positive rate (TPR) and false-positive rate (FPR) of discovering true causes.
\begin{figure*}[ht]
    \centering
    \subfigure[CSL algorithms in an ensemble have higher average agreement on true causes.]{
        \includegraphics[width=0.44\linewidth,keepaspectratio]{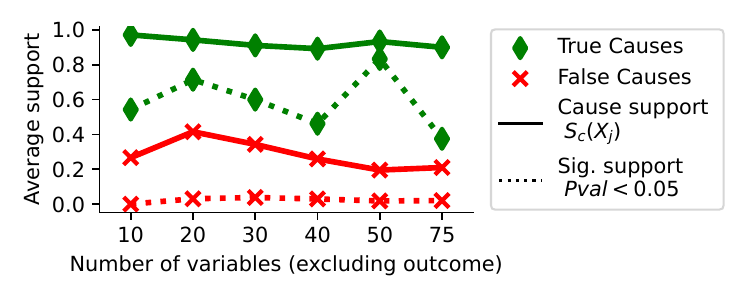}
        \label{fig:semisupport}
    }
    \subfigure[A majority voting ensemble and individual algorithms for different data generation, graph topology, and edge density in terms of 1-F1 score.]{
    \includegraphics[width=0.52\linewidth]{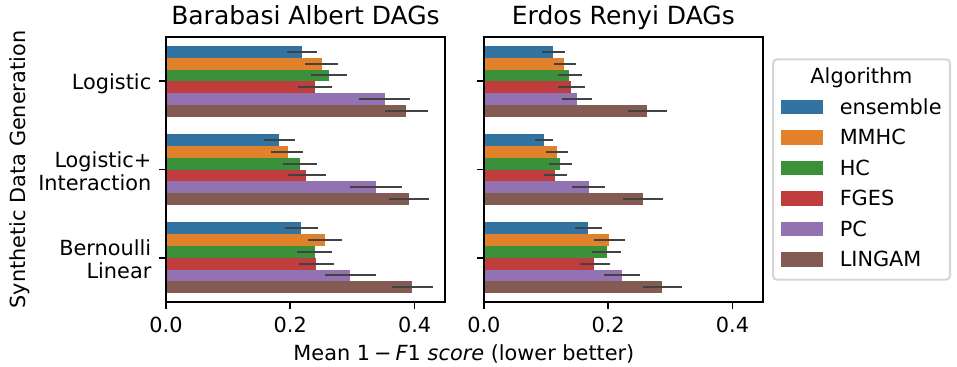}
     \label{fig:sensitivity}
    }
    \subfigure[A majority voting ensemble and individual algorithms in terms of recall and F1 scores.]{
        \includegraphics[width=0.88\linewidth,keepaspectratio]{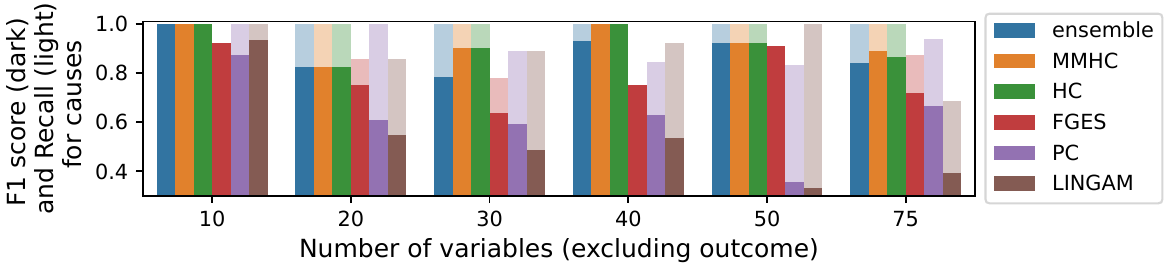}
    \label{fig:semiassum}
    }
    \vspace{-1em}
    \caption{Evaluation of the ensemble approach with synthetic (b) and semi-synthetic (a and c) data.}
    \vspace{-1em}
\end{figure*}
% \begin{figure}
%     \centering
%     \includegraphics[height=0.2\linewidth]{images/ba_ancEval_all.pdf}
%     \caption{Caption}
%     \label{fig:sensitivity}
% \end{figure}

\emph{The ensemble approach can rank potential causal hypotheses.}
Figure \ref{fig:semisupport} shows that the CSL algorithms in an ensemble, for semi-synthetic data, have more than 90\% average agreement (i.e., cause support) for true causes, whereas the average agreement for false causes is far less. This indicates that CSL algorithms, despite diverse approaches and assumptions, tend to agree more on true causes than on false causes. Similarly, as seen in Figure \ref{fig:semisupport}, the estimated total effects for false-positive causes are mostly statistically insignificant ($p \ge 0.05$). Synthetic experiments included in the Appendix yield a similar result. Therefore, the cause support metric, the magnitude of the total effects, and the confidence scores with p-values obtained from the ensemble help rank the causal hypotheses.

\emph{A simple majority voting ensemble to determine the causes offers robustness to data generation mechanisms, network topology, number of nodes, and density of edges.} The performance of a majority voting ensemble and individual algorithms for identifying causes of $Y$ is summarized in Figure \ref{fig:sensitivity} in terms of the $(1-F1~score)$ metric for three synthetic data generation mechanisms and two DAG generation mechanisms. For each DAG type and data generation type, the ensemble approach outperforms any individual algorithm on average across DAGs with varying numbers of nodes and edge densities.  The results show that the ensemble provides increased robustness over committing to a single algorithm. The Appendix explains the detailed experimental setup and additional results consistent with the above observation.

%The estimated ATE for true-positive causes may occasionally be statistically insignificant due to invalid estimated adjustment sets or unfaithful data distributions. These results indicate that the ensemble approach enriches potential causal hypotheses with the help of cause support and significance support metrics.

% \emph{The ensemble approach adds robustness to heterogeneity discovery that is more sensitive to causal structure misspecifications}. To test the ensemble approach for heterogeneity discovery described in Section \ref{sec:discoverHte}, we evaluate the discovery of the ground truth effect modifiers for each true cause and outcome. Figure \ref{fig:semisupport} shows higher agreement on true interactions terms, i.e. $(X^j \times Z)|X^j \in anc(Y, G_T) \land  \text{Eq. \ref{eq:hte} holds}$, than false interactions terms absent in the ground truth model.

\emph{A simple majority voting ensemble to determine the causes favors high recall while maintaining fair precision.}
Figure \ref{fig:semiassum} shows the performance of individual algorithms and the ensemble in terms of recall and F1 scores for the semi-synthetic data generation. We notice, as expected, MMHC and HC with similar methodology or assumptions with the underlying data generation process have good performances, while other algorithms with different assumptions and methodologies have slightly poorer performances. LINGAM with incompatible assumptions has a lower F1 score. The ensemble is able to maintain a high recall with a decent F1 score. 
% These observations are consistent across several synthetic experiments included in Appendix  due to limited space.
% The assumptions satisfied by the real-world data are almost always untestable, and selecting a single causal model is difficult.
We note that majority voting may perform poorly when the performance of each base algorithm is poor, for example, due to a limited sample size. Therefore, we prefer ranking the causes utilizing confidence measures as demonstrated with real-world results.
The results of synthetic and semi-synthetic experiments show that the ensemble approach is robust in discovering potential causes with a measure of confidence.
% \fixme{Add some results on deep learning methods.}
% \vspace{-2em}
\subsection{\rev{Expert Clinician Validation}}
\rev{For a subset of findings in the repeated ED visit analysis for the diabetic population, we seek validation from expert clinicians. We extract causal hypotheses from the cause-and-effect relationships among the ancestors of repeat ED visits identified by the MMHC algorithm (chosen for its reliable performance on synthetic data and sparse output). Due to the limited time of the clinicians, we focused on $30$ discovered hypotheses. Some of the related causal hypotheses were merged into a single question asked to the clinicians. Four independent expert clinicians judged causal-and-effect hypotheses and responded 1) whether the hypothesis was likely true, 2) whether the reverse causal direction was more likely, and 3) neither. 
The detailed instructions provided to the clinicians about the evaluation format, along with 30 questions, are available as supplementary materials. There was a majority agreement among three or more clinicians for $83\% (25/30)$ of the questions, with full agreement on $30\% (9/30)$ of the questions. The overall agreement among clinicians was assessed using commonly used metrics: Krippendorff's $\alpha = 0.309$ and Fleiss' $\kappa = 0.312$, indicating fair agreement.}

\rev{For the evaluated causal relationships identified by the MMHC algorithm, there were 17 (57\%) true positive causal-and-effect relationships (i.e., majority of evaluators agreed with the algorithm), 10 (33\%) false positive causal-and-effect relationships (i.e., majority of evaluators did not agree with the algorithm and evaluated as reverse or neither), and 3 (10\%) relationships were undecided due to disagreement among evaluators (i.e., half agreed with the algorithm). Of the 10 false-positive relationships, the majority of evaluators agreed that 4 were reversed in causal direction and 5 were annotated as "neither", with the remaining one being annotated as reverse by two out of four evaluators.
We compare the agreement between CSL algorithms in the ensemble for confirmed true causal relationships and confirmed non-existent relationships. We check whether the other two algorithms have causal pathways that match the evaluated cause-and-effect relationships to obtain the ensemble's cause support metric. The average cause support (agreement between algorithms) was 69.57\% for confirmed true causal-and-effect relationships and 37.14\% for confirmed non-existent causal relationships. This evaluation aligns with the simulated experiments, in which the ensemble has higher agreement between algorithms for true causes and lower agreement for the false causes. Interestingly, gradient-based methods GOLEM and GraNDAG only identified two confirmed causal-relationships and rejected all but two non-existent relationships.
}

\rev{The majority of clinicians agree with the findings that diagnosis codes related to the general medical exam (Z000) and the encounter for immunization (Z23) are preventive factors of repeat ED visits. However, unlike CSL algorithms in the heterogeneous causal discovery framework, all clinicians agreed that the diagnosis code for mixed hyperlipidemia (E785) does NOT reduce the likelihood of repeat ED visits. A diagnosis code may also signal that a patient has been evaluated and made aware of a risk state, not just that the disease exists. Clinicians should examine the hypothesis that this awareness triggers risk-reducing actions (e.g., glucose or lipid monitoring and lifestyle changes), which could lower future complications despite the underlying morbidity.}

%% file: 7-discussion.tex
\section{Discussion}\label{sec:discussion}

\subsection{\rev{Utility to Clinical Investigators}}
\rev{Our framework generates clinically interpretable causal hypotheses (candidate drivers, preventive factors, and effect modifiers) that investigators can subsequently evaluate using established study designs. The workflow is intended to complement rigorous causal inference rather than replace it, and it can be applied beyond healthcare utilization outcomes to other clinical endpoints in EHR or claims data.}

\rev{\textbf{From hypothesis generation to study planning}. A practical advantage of our framework is that it produces a \textbf{multiset of effect-size estimates} for candidate interventions under different data-generating assumptions. Effect size is a key input to power analysis and sample-size calculations, which determine how many participants are needed to detect an effect with an acceptable Type II error (i.e., the chance of missing a true effect). These considerations are routinely required in clinical study protocols.  This is particularly valuable when recruitment is difficult and costs scale with cohort size, because an initial effect estimate helps investigators judge feasibility and prioritize which hypotheses merit expensive follow-up studies.}

\rev{\textbf{Targeting the right patients}. Clinical trials and prospective observational studies increasingly emphasize that average effects can mask meaningful heterogeneity in treatment effects. Some subgroups benefit more, others less, and some may experience net harm. Our framework identifies clinically meaningful effect modifiers informing subpopulations with differential effects. For example, patients with a history of seizures/convulsions or sepsis were identified to have attenuated effects of preventive interventions. Such insights can inform inclusion or exclusion criteria in study designs, support prespecified subgroup analyses, and motivate enrichment designs that prioritize likely responders.}

\rev{\textbf{Applications beyond healthcare utilization}. Our framework can be \textbf{adapted} to other repeated undesirable health outcomes beyond healthcare utilization (e.g., recurrent myocardial infarction or repeat revascularization procedure). For instance, in a revascularization procedure, doctors restore blood flow to the heart using percutaneous coronary intervention (PCI) with stent placement to keep the artery open or coronary artery bypass grafting (CABG) to reroute the blood flow around a blocked artery. In this context, our framework could generate candidate drivers of repeat revascularization risk (e.g., comorbidity profiles, post-procedural complications, and medication patterns), quantify their estimated effects, and provide actionable insights for designing targeted prospective observational studies or randomized trials.}

\subsection{\rev{Assumptions, Limitations, and Future Research Directions}}

\rev{\textbf{Latent confounding and model misspecification}. A central assumption of causal discovery and effect estimation from observational EHR or claims data is the absence of latent confounding. However, due to unobservable factors or incomplete measurements, latent confounding can occur, potentially leading to inaccurate causal relationships, biased treatment effects, and spurious effect modifiers. The use of high-dimensional features and causal inference with observational data can sometimes mitigate the effects of unobserved confounding. For example, expert clinician validation revealed that 21 out of 27 relationships had correct causal adjacency. Future work should incorporate clinician-guided variable augmentation and incorporate causal discovery methods that explicitly allow latent confounders (e.g., FCI~\cite{spirtes-book00}). Similarly, we rely on a simple but interpretable linear models with interactions for estimating causal effects by utilizing causal model-informed discovery of potential effect modifiers. This approach is common in healthcare settings and suitable for mostly binary data with a decent adjustment set like ours. However, to mitigate the risk of model misspecification and account for high-dimensional adjustment set, causal machine learning~\cite{athey-pnas16,chernozhukov-tej18} approaches can be used.}

\rev{\textbf{Feature representation, interpretibility, and consistency assumption}. By using the presence or absence of historical diagnosis codes as features, the resulting interventions are interpretable and mostly well-defined. However, this simplification implicitly assumes consistency (i.e., a single version of the exposure), even though in practice a binary indicator can bundle clinically different situations (e.g., one visit versus regular follow-up). In real-world applications, more clinically meaningful representations (e.g., intensity, recency, or severity proxies) may be needed to preserve well-defined interventions. Furthermore, it remains an open question how to incorporate rich clinical embeddings from foundational models~\cite{wornow-neurips23} into causal discovery and effect estimation without losing interpretability. Relatedly, we use a summary graph and limited longitudinal information to impose structural constraints and help orient undirected edges. In our application, the timing of disease onset and the timing of diagnosis documentation can differ (e.g., delayed recognition or care-seeking preferences). Therefore, we treat temporal cues as weak constraints rather than definitive evidence of causal direction. However, temporal causal discovery methods~\cite{cai-tnnls22,gong-arxiv23} that explicitly model event timing may be useful when timestamps closely reflect causal ordering.}

\rev{\textbf{Faithfulness, selection, and censoring}. Most causal discovery methods assume faithfulness, meaning that the conditional independencies observed in the data reflect the relationships entailed by the underlying causal graph. In finite samples, this assumption can be effectively violated because conditional-independence tests become noisy. As illustrated with the simulated data in the Appendix Table 13, %\ref{samples_ba_linear_ancEval}, 
extremely small sample sizes degrade all algorithms and can also reduce the ensemble's benefit. Our primary cohorts are sufficiently large, but the effective sample sizes for EHRSHOT endpoints are smaller, so we use EHRSHOT mainly to demonstrate portability and to assess whether broad themes persist across populations, rather than to claim definitive general-population drivers of repeated healthcare utilization. Our cohorts are also defined conditional on an index event (e.g., an initial ED visit or discharge), which could introduce selection bias if factors influencing entry into the cohort induce spurious associations. In both claims and EHR settings, loss to follow-up (e.g., patients leaving the system, death, or missing encounters) can bias estimates if censoring is not random. In our primary analyses, we include patients with claims or health records after the time horizon if they do not encounter an event. Future work should consider survival models to account for non-random censoring due to deaths and loss to follow-up.}

%% file: 8-conclusion.tex
% \vspace{-1em}
\section{Conclusion}\label{sec:conclusion}
In this paper, we investigate the problem of discovering causes and effect modifiers of an outcome using observational data. We focus on practical scenarios where there is uncertainty in cause-and-effect hypotheses generated by CSL algorithms. We propose a framework that incorporates an ensemble of CSL algorithms, effect modifiers discovery, and heterogeneous effect estimation to output causes and effect modifiers of the outcome consistent across multiple causal models. Using synthetic and semi-synthetic experiments, we show the benefits of using an ensemble of CSL algorithms for discovery tasks. We demonstrate the promise of our approach with real-world use cases of discovering causes and sources of heterogeneity of repeat ER visits for diabetic patients and hospital readmission for ICU patients. The evaluation of the results utilizing published research literature shows our framework automatically generates causal hypotheses consistent with known knowledge. We demonstrate the portability of our methods in modern electronic health records using the EHRSHOT dataset. Evaluation by expert clinicians on a subset of findings show reliability of our framework for automatic causal hypothesis generation. We discuss the downstream utility for clinical investigators and future research directions.
% The automatically generated causal hypotheses of preventive and risk factors, as well as effect modifiers of repeat ER visits, could be further investigated by subject matter experts.

\section*{Acknowledgments}
We would like to thank Dr. Neomi Shah for her assistance in recruiting clinicians from Mount Sinai for expert review. We also extend our gratitude to all four clinicians who volunteered their time for completing our survey.

%% file: 9-appendix.tex
\onecolumn
% \twocolumn
\section*{Appendix}
Here we provide the following supplementary materials for the main paper:
\begin{enumerate}
    \item Description of causal structure learning algorithms,
    \item Description and results of simulated data experiments,
    \item  Details, including pseudocode, for orienting undirected edges of a DAG's equivalence class,
    \item Additional details of the experimental setup and hyperparameters.
    % \item steps to summarize multi-set of causal effects for top causes on the repeated undesirable outcome, and
\end{enumerate}

\subsection*{Simulated Experiments}
Here, we provide additional details on simulated experiments including random graph and data generation as well as experimental results supporting the results presented in the main paper.
% The advantage of using simulated data is that it contains ground truth.
A simulated experimental setup is necessary for the evaluation of causal discovery methods because there is a lack of ground truth in real-world datasets. Synthetic data provides flexibility to test our hypothesis. However, using purely synthetic data for evaluating causal models may have limitations like high “researcher degrees-of-freedom” and lack of standardization~\cite{gentzel-nips19}.
%Moreover, synthetic data generation representing the phenomena of repeated undesirable outcomes is non-trivial.
A semi-synthetic data generation strategy, that utilizes features of real-world data, helps in mitigating some of these limitations.
Therefore, we employ both synthetic and semi-synthetic data generation methods to evaluate the benefits of our causal structural learning ensemble over using a single CSL algorithm. Synthetic data generation representing the phenomena of repeated undesirable outcomes is non-trivial. Instead, we focus on simulated experiments with IID data obtained with Bag-of-Events (BOE) representation.\\
\textbf{Ground truth causal graph generation}

\textbf{\textit{Synthetic}}. We want to test how sensitive CSL algorithms are to structural properties like the number of nodes, the sparsity of edges, and the network topology in discovering the causes of a sink outcome node. For this reason, we generate ground-truth directed acyclic graphs (DAGs) according to Erdos-Renyi (ER) and Barabasi-Albert (BA) random graph models by varying the number of nodes ($n$) and connectivity parameters. The undirected ER graph is controlled by the edge probability parameter $p$, while the BA graph is controlled by the preferential attachment parameter $m$, referred to as the sparsity parameter ($sp_{ba}$). For consistency, we define the sparsity parameter $sp_{er} | p = \frac{2 \times n \times sp_{er}}{n\times (n-1)}$ for the ER graph as well indicating average degree of graph.

Next, we orient the undirected edges of the generated ER and BA graphs according to a random topological order of nodes. We also add a sink outcome node to the oriented graph. For the ER graph, k nodes are connected to the outcome node such that the average degree $sp_{er}$ is maintained. For the BA graph, we calculate the uniform edge probability for the existing graph and then randomly connect existing nodes to the outcome node according to the edge probability. This process gives us random DAGs with a sink outcome node.

\textbf{\textit{Semi-Synthetic}}. In addition to randomly generated graphs, we use real-world data to generate ground truth causal graphs and learn their SCM parameters. We use a base algorithm to learn a causal graph and its parameters which we consider as the ground truth. We present analysis with MMHC as base algorithm but we get similar trends for other base algorithms. The causal model is learned from the real-world insurance claims data. The data is preprocessed with $\tau=12$ months as time interval for repeated undesirable outcome. Top $N \in \{10,20,30,40,50,75\}$ most frequent events along with the outcome are subset from the aggregated Bag-of-events (BOE) data representation based on frequency of diagnosis codes for generating ground-truth graphs of different sizes. The top N include an out-of-vocabulary (OOV) variable indicating presence of other less frequent events. This data is used to learn a causal model $G_T(\mathbf{V_T}, \mathbf{E_T})$. If any edges in the causal model are unoriented, we use orientation support, defined in the next section, to orient these edges.\\ 
 %The ground truth causal graph and parameters generated by the above procedure are more standardized and realistic than randomly generated ones.
\textbf{Data generation}

\textbf{\textit{Synthetic}}. We consider three modes of data generation for synthetic experiments, focusing on binary data, with the following functional form:
\begin{enumerate}
    \item \textit{Logistic (L)}. $X=Bernoulli(sigmoid(b + \mathbf{W}\times pa(X,G)))$, where $G$ is ground truth graph, $b \in \mathbb{R}$, and $\mathbf{W} \in \mathbb{R}^{L}$ with $L=cardinality(pa(X,G))$. 
    \item \textit{Logistic + Interaction (LL)}. $X=Bernoulli(sigmoid(b + \mathbf{W}\times pa(X,G)) + \mathbf{I}\times combination(pa(X,G), 2)))$, where $\mathbf{I} \in \mathbb{R}^{\binom{L}{2}}$, $b \in \mathbb{R}$, and $\mathbf{W} \in \mathbb{R}^{L}$.
    \item \textit{Bernoulli Linear (BL)}. $X=Bernoulli(b+\mathbf{W}\times pa(X,G))$, where $b \in (0,1)$, $\mathbf{W} \in (-1,1)^L$, and $0 \le b+\mathbf{W}\times pa(X,G) \le 1$.
\end{enumerate}

\textbf{\textit{Semi-Synthetic}}.
From the ground-truth causal graph. we estimate the SCM parameters as conditional probability tables (CPT) $p(V_i|pa(V_i, G_T))$. Using the SCM parameters and the topological ordering of the ground truth causal DAG, we sample synthetic data in the BOE representation.\\
\textbf{Evaluation metrics}

% We run an ensemble of five CSL algorithms -- MMHC, HC, FGES, PC, and LINGAM -- on the semi-synthetic data. 
To assess the performance of the ensemble approach in refining potential causal factors, we estimate the average causal support for the true causes ($TP(S_c)$) and the false causes ($FP(S_c)$) defined as
% \begin{equation}
    $TP(S_c) = \frac{\sum_{X^j \in anc(Y, G_T)}S_c(X^j)}{n(anc(Y, G_T))}, \text{and}$
    $FP(S_c) = \frac{\sum_{X^j \notin anc(Y, G_T)}S_c(X^j)}{n(G_T) - n(anc(Y, G_T))}$,
% \end{equation}
where $Y$ is the outcome and $S_c(X^j)$ is the cause support for event $X^j$. The $TP(S_c)$ and $FP(S_c)$ measures indicate the agreement among multiple CSL algorithms for true causes and false causes, respectively. In addition to the structural cause support, we investigate significance support by checking the fraction of algorithms with statistically significant ($pvalue < 0.05$) total effects for the identified causes. The significance support gives a measure of confidence for discovered causes. Moreover, we compare individual algorithms and the ensemble in terms of precision, recall, and F1 scores of discovering true causes.\\
\textbf{Simulated experiment results}

The main paper presented results from the semi-synthetic experimental setup. Here, we describe the evaluation results for the following synthetic experimental setups.

\begin{figure}[t]
        \centering
        \subfigure[ER random DAGs]{\includegraphics[width=0.45\linewidth,keepaspectratio]{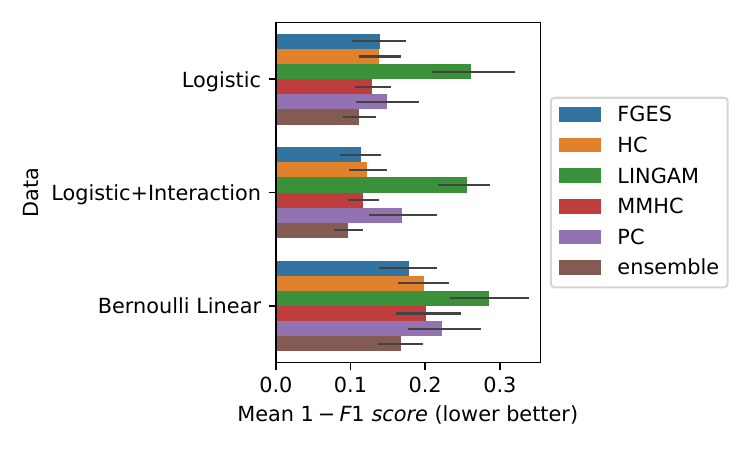}
    % \caption{.}
    \label{fig:er_diff}}
    \hfill
\subfigure[BA random DAGs]{
\includegraphics[width=0.45\linewidth,keepaspectratio]{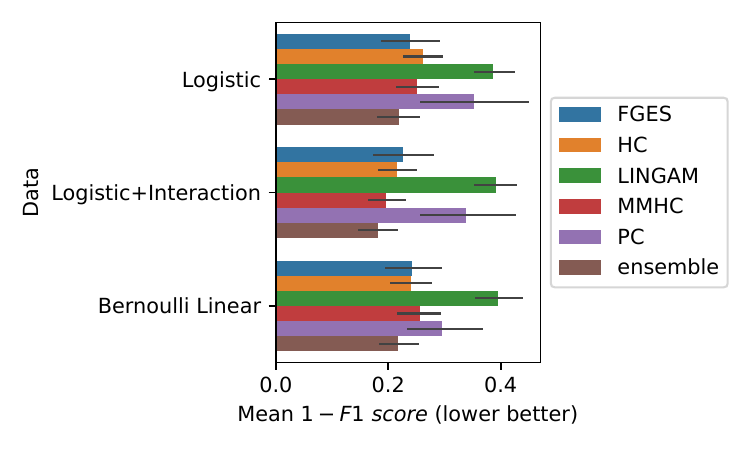}
        \label{fig:ba_diff}}
    \caption{Summarizing Tables 7 to 12: Performance of individual algorithms and ensemble in terms of mean 1-F1 score (lower better). Ensemble reduces the sensitivity to variation in structural properties like the number of nodes, the sparsity of edges, and network topology. An ensemble provides robustness compared to selecting a single algorithm.}
    \label{fig:results_diff_st}
\end{figure}

\textit{\textbf{Sensitivity to structural properties like the number of nodes, the sparsity of edges, and the network topology}}. In this setup, we randomly generate DAGs with ER and BA topologies by varying nodes (excluding sink outcome) $n \in \{10, 20, 30, 40\}$ as well as sparsity parameters $sp_{er} \in \{0.8, 1.0, 1.5, 2.0\}$ and $sp_{ba} \in \{1, 2, 3, 4\}$. For each combination of the number of nodes and sparsity parameters, we generate $5$ random DAGs. Then, we generate data according to logistic (L), logistic + interaction (LL), and Bernoulli linear (BL) functional forms $3$ times each with different parameters for each DAG. We run an ensemble of five CSL algorithms: PC, FGES, HC, MMHC, and LINGAM. The assumptions of first four algorithms are similar but they differ in their methodology and hyperparameters (describe later). LINGAM, although incompatible with binary data, is included to test the robustness of ensemble toward noisy base learner. PC and FGES may return CPDAGs while MMHC, HC, and LINGAM return DAGs. We run bootstrapped version of MMHC and HC algorithms for 20 runs and leave edges undirected if they support each orientation 50\% of the time. The evaluation results consider an $anc(Y, G_{est})$ query that returns all ancestors with a directed path to the outcome node $Y$, ignoring undirected edges in the estimated graph.

Tables 1 to 6 show the average causal support (agreement) of the ensemble for true causes and false causes of the sink outcome node. These results are consistent with Figure 6 in the main paper, where the algorithms in the ensemble have higher agreement (support) on true causes and lower support for false causes. The significance support shows that it can be used to rank the causes for higher precision.

Tables 7 to 12 show the performance of the ensemble compared to each algorithm in terms of F1 score for different settings of structural parameters. It is observed that the performance of algorithms varies according to the structural properties and the ground truth structure, although most of the assumptions are satisfied and the sample size is decent ($n\times 1000$). It is revealed that the ensemble has the least deviation from the maximum F1 score. Figure 2 summarizes Tables 7 to 12, showing the ensemble has the best performance on average compared to the individual base algorithms. This indicates an ensemble provides robustness compared to selecting a single algorithm for practical settings with uncertain ground-truth structural properties. We notice the CSL algorithms have relatively better performance for ER model compared to BA model with preferential attachment mechanism. In both topologies, the ensemble improves the performance on average.

\input{images/synthetic_exps/er_logistic_ancEval_support}
\input{images/synthetic_exps/er_logistic_int_ancEval_support}
\input{images/synthetic_exps/er_linear_ancEval_support}

\input{images/synthetic_exps/ba_logistic_ancEval_support}
\input{images/synthetic_exps/ba_logistic_int_ancEval_support}
\input{images/synthetic_exps/ba_linear_ancEval_support}

\input{images/synthetic_exps/er_logistic_ancEval}
\input{images/synthetic_exps/er_logistic_int_ancEval}
\input{images/synthetic_exps/er_linear_ancEval}

\input{images/synthetic_exps/ba_logistic_ancEval}
\clearpage
\input{images/synthetic_exps/ba_logistic_int_ancEval}
\input{images/synthetic_exps/ba_linear_ancEval}

\begin{figure}[th]
        \centering
\subfigure[BA random DAGs]{
         \includegraphics[width=0.45\linewidth,keepaspectratio]{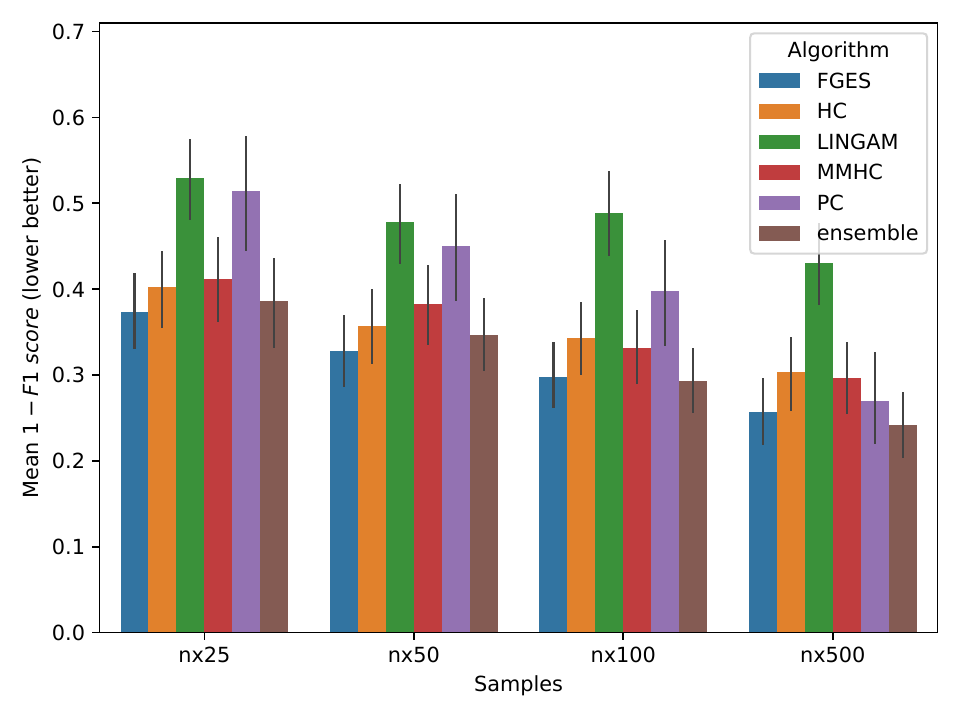}
        \label{fig:samples_diff}
        }
    \hfill
       \subfigure[ER random DAGs]{    \includegraphics[width=0.45\linewidth,keepaspectratio]{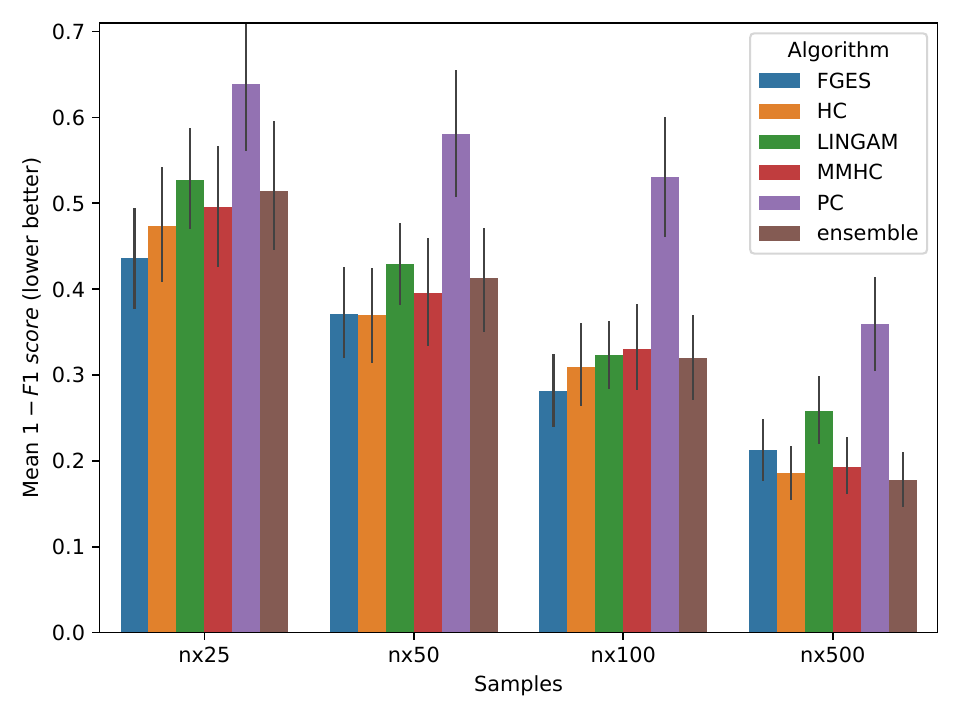}
    \label{fig:er_diff}}
    \caption{Sensitivity of CSL algorithms to potential unfaithfulness due to limited sample size (n=30 nodes, sparsity=2). Ranking causes according to cause support, magnitude of effects, and significance support should be prioritized over a (majority) voting ensemble for robustness.}
    \label{fig:samples_diff_st}
\end{figure}

\textit{\textbf{Sensitivity to effect size and sample sizes (potential unfaithfulness)}}.
The constraints in the Bernoulli linear data generation mechanism may produce weights of small magnitude (i.e., low effect size). Figure \ref{fig:results_diff_st} shows CSL algorithms have relatively poor performance for the ER model with Bernoulli linear data generation. However, for the BA model, the performance is similar for all three data generation options.
The experimental setup used to assess whether CSL algorithms are sensitive to probable unfaithfulness caused by a small number of samples is described next. We generate 100 random DAGs consisting of $n = 30$ nodes with an edge sparsity of $2$ for both the ER and BA models. Then, for each DAG, we evaluate the ensemble of five algorithms described above with sample sizes of $n\times 25$, $n \times 50$, $n \times 100$, and $n \times 500$. Table 13 and Figure \ref{fig:samples_diff_st} summarize the results of this experiment. For limited samples, the CSL algorithms are more likely to miss the true causes, and the average support for the true causes is lower. Thus, a majority voting ensemble suffers due to limited samples. A ranking scheme utilizing cause support, magnitude of causal effects, and significance support may be more preferable than a threshold voting scheme. The results for real-world applications in the main paper follow this ranking scheme.

\input{images/synthetic_exps/samples_er_ba_support}

\textbf{Ranking causes}.
Here, we present the procedure for summarizing risk factors or preventive factors of a repeated undesirable outcome, as shown in Figure \ref{fig:results_atereadmit}. Our framework outputs a list of causes with two properties associated with each cause: cause support ($S_c(X^j)$) and multi-set of total causal effects $\mathbf{T}(X^j):=\{\forall{k \in \{1,...,K\}}, X^j_{CATE}(\mathbf{Z}=\emptyset) \mbox{ if } (X^j, G_k) \in \mathbf{R} \mbox{ else 0}\}$. For generating top risk factors, the causes are ranked in the descending order of cause support $S_c(X^j)$ as a primary sort key and the descending order of maximum causal effect in a multi-set, i.e. $max(\mathbf{T}(X^j))$, as a secondary sort key. The top $N$ risk factors are displayed according to descending order of $max(\mathbf{T}(X^j))$. The stacked bars show the multi-set effects with lower values overlapping the higher values. 
For generating top preventive factors, the secondary sort is done in the ascending order of minimum causal effect in a multi-set, i.e. $min(\mathbf{T}(X^j))$. The top $N$ preventive factors are displayed according to the ascending order of $min(\mathbf{T}(X^j))$. This procedure generalizes to the effect modifier analysis, as depicted in Figure 5, where heterogeneous causal effects are considered instead of total causal effects.

\subsection*{Orienting Undirected Edges}
\newchange{\textbf{Orientation support}.
The undirected edges due to the equivalence class of a DAG are oriented utilizing the temporal order of events prominent across population $\mathbf{P}$ in the event-log data model. We define \textit{orientation support} for an edge direction from {\small $V_j$} to {\small $V_k$} {\small $V_j \rightarrow V_k$} as {\small
% \begin{equation}\label{eq:orient_support}
$S_o(j,k)=\frac{1}{n(\mathbf{D})}\sum_{i\in \mathbf{D}}{\llbracket min(t|e^j_{i,t}=1) < min(t|e^k_{i,t}=1) \rrbracket}$},
% \end{equation}}
where {\small $\mathbf{D}\subseteq \mathbf{P}:=\{E_i|\exists{p}, e^j_{i,p}=1 \land \exists{q}, e^k_{i,q}=1\}$} is subpopulation with both events $e^j$ and $e^k$ in a timeline $E_i$ and $\llbracket.\rrbracket$ is an indicator function. $S_o(j,k)$ gives the fraction of units with the parent event $e^j$ before the child event $e^k$ for a given edge orientation. We calculate $S_o(j,k)$ for all the possible orientations of undirected edges, sort the edges from highest to lowest support, and orient the edges by handling the constraints of no cycles and no new v-structures to obtain one DAG per CSL algorithm.}
% The Appendix includes pseudocode with additional details.

Algorithm \ref{alg:orient} shows our greedy approach of orienting undirected edges for any causal structure learning (CSL) algorithm that outputs an equivalence class of a DAG. The input to Algorithm \ref{alg:orient} is a set of edges $\mathbf{E}$ where direction can be a null value for undirected edges and the output is a set of edges $\mathbf{D}$ without null direction. First, we calculate $S_o(j,k)$, defined in the Methodology Section, for all the possible orientations of undirected edges (lines 5-6). The undirected and directed edges are stored to $\mathbf{U}$ and $\mathbf{D}$ respectively (line 7-9).
Second, we sort the edge orientations from highest to lowest support (line 12). Finally, we orient the edges by handling the constraints of no cycles and no new unshielded v-structures to obtain one DAG per CSL algorithm (lines 13-26).
{\small
\begin{algorithm}[ht]
\caption{Orienting undirected edges. \\Input: $\mathbf{E}$, Output: $\mathbf{D}$} \label{alg:orient}
\begin{algorithmic}[1]
\State $\mathbf{S} \gets \{\}$
\State $\mathbf{D} \gets \{\}$
\State $\mathbf{U} \gets \{\}$
\For{$(V_j, direction, V_k) \in \mathbf{E}$}
\If{$direction = \emptyset$}
\State $\mathbf{S} \gets \mathbf{S} \cup \{(V_j, V_k, S_o(j,k)), (V_k, V_j, S_o(k,j)\}$
\State $\mathbf{U} \gets \mathbf{U} \cup \{(V_j - V_k)\}$
\Else
\State $\mathbf{D} \gets \mathbf{D} \cup \{(V_j~direction~V_k)\}$
\EndIf
\EndFor
\State $\mathbf{S} \gets sort\_desc(\mathbf{S}, key=S_o(j,k)\in (V_j, V_k, S_o(j,k))$
\For{$(V_j, V_k, S_o(j,k)) \in \mathbf{S}$}
\If{$(V_k \rightarrow V_j) \in \mathbf{D}$}
\State \textbf{goto} continue
\EndIf
\State $G \gets graph(\mathbf{D} \cup \mathbf{U})$
\State $\mathbf{P} \gets pa(V_k, G)$
\If{$n(\mathbf{P})>0 \land \exists{V_i \in \mathbf{P}}, \lnot adj(V_j, V_i, G)$}
% \State $\mathbf{C} \gets \mathbf{C} \cup (V_j \rightarrow V_k)$
% \If{$(V_k \rightarrow V_j) \in \mathbf{C}$}
% \State $\mathbf{E} \gets \mathbf{E}\setminus \{(V_j \rightarrow V_k), (V_k \rightarrow V_j)\}$
% \EndIf
\State \textbf{goto} continue
\EndIf
\State $\mathbf{N} = \{(V_j \rightarrow V_k)\} \cup Meek95((V_j \rightarrow V_k), G)$
\State $\mathbf{D} \gets \mathbf{D} \cup \mathbf{N}$
\State $\mathbf{U} \gets \mathbf{U} \setminus undirected(\mathbf{N})$
\State \textit{continue:}
\EndFor
\end{algorithmic}
\end{algorithm}
}

We iterate through all the possible ordered orientations and add new edge orientations to the set of edges (lines 22-23) if there are no conflicts. Lines 14 to 16 check the cycle due to the edges already oriented in the reverse direction in prior iterations. 
A new unshielded v-structure is introduced if, for a proposed orientation, the child node already has other parents that are not adjacent to the proposed parent in the current graph $G$ (line 17-19). The addition of such an orientation is skipped (line 20). 

The addition of a new orientation is followed by orientation of the graph $G$ according to the rules by \citet{meek-uai95} (line 22). The implementation of orientation using \citet{meek-uai95} rules is available via \texttt{pcalg}\footnote{https://cran.r-project.org/web/packages/pcalg/index.html} package~\cite{perkovic-uai17}. The resulting output DAG contains all the edges in $\mathbf{D}$. Algorithm \ref{alg:orient} uses the orientation order as heuristics to orient the undirected edges of an equivalence class, and ensures soundness by not introducing any new edge, cycle, or unshielded collider, as well as not deleting any undirected edge. 
While the above algorithm ensures scalability, a brute force implementation by enumerating all valid orientations and selecting orientation with maximum orientation support is more suitable for a few undirected edges.
% \fixme{Soundness: Cycles and undirected edge deletion issues?}
\subsection*{Experimental Setup}
Here, we first specify our computing infrastructure along with additional software packages used in the experiments. Then, we describe the hyperparameters of the five causal structure learning algorithms considered in the experiments: MMHC, HC, FGES, PC, and LINGAM.

\textbf{Computing infrastructure}.
We run our experiments on a device with 64-cores CPU (AMD Opteron(TM) Processor 6274 model), 118 GB RAM, 4 TB HDD, and Ubuntu 18.04 operating system. We use \texttt{Anaconda}\footnote{https://www.anaconda.com/} for Python package and virtual environment management. We use the \texttt{bnlearn} package with R version 4.0.3 for implementing HC and MMHC algorithms. For the implementation of LINGAM algorithm, we use R's \texttt{pcalg} package. The FGES and PC algorithms are implemented using \texttt{py-causal} package which is a Python wrapper for Java-based \texttt{Tetrad}\footnote{http://cmu-phil.github.io/tetrad/manual/} package for causal structure learning. We use another virtual environment with Python 3.7. 

We utilize Python's \texttt{networkx} package for graph processing and \texttt{statsmodels} package for regression and non-parametric estimation of causal effects. Python packages \texttt{pandas} and \texttt{numpy} are used for data generation, cleaning and processing, while \texttt{matplotlib} and \texttt{seaborn} packages are used for visualization. Our code and semi-synthetic dataset will be publicly available after publication.

\textbf{Hyperparameters}.
We first describe the hyperparameters for the discrete version of the FGES algorithm available with \texttt{py-causal} package. FGES uses a BDeu (Bayesian Dirichlet likelihood equivalence and uniform)~\cite{heckerman-jm95} as a scoring function that ensures the same dependence and independence relations get the same score. Other parameters and their values are as follows.
\begin{itemize}
    \item \texttt{priorKnowledge}=\{\texttt{forbiddirect}:$\{\forall{j\in\{1,...,M\}},\\(Y \rightarrow X^j)\}$\}, a dictionary with values containing a set of restricted edges according to the structural constraints defined in the Methodology Section.
    \item \texttt{dataType}=\texttt{"discrete"}, specifies we have discrete data.
    \item \texttt{faithfulnessAssumed}=\texttt{False}, specifies data distribution may violate the faithfulness assumption, and a robust implementation should be used.
    \item \texttt{maxDegree} = $-1$, specifies no assumptions on the sparsity of the underlying causal graph and no restriction on the maximum degree of the output graph.
    \item \texttt{symmetricFirstStep}=\texttt{True}, a setting recommended in the documentation for discrete causal structure search.
\end{itemize}

PC algorithm uses \texttt{priorKnowledge} and \texttt{dataType} similar to the FGES algorithm. We set the following hyperparameters and leave all other parameters as default.
\begin{itemize}
    \item \texttt{testId}=\texttt{"chi-square-test"}, conditional independence test for categorical variables.
    \item \texttt{concurrentFAS}=\texttt{True}, use parallized fast adjacency search.
    \item \texttt{depth}=\texttt{5}, maximum number of conditioning set (ensuring algorithm runs in reasonable time).
    \item \texttt{conflictRule}=\texttt{2}, if conflicts arises in collider discovery, then orient conflict as bidirected edge like FCI algorithm.
    \item \texttt{colliderDiscoveryRule}=\texttt{3}, collider orientation by choosing the separation-sets with the maximum p-value. Robust for small sample sizes.
\end{itemize}

For the HC algorithm, we set the following hyperparameters and leave all other parameters as default. 
\begin{itemize}
    \item \texttt{score}=\texttt{"bic"} (default), specifies BIC (Bayesian Information Criterion)~\cite{chickering-uai95} scoring function.
    \item \texttt{maxp}=\texttt{Inf} (default), specifies the maximum number of parents in the output graph and is equivalent to \texttt{maxDegree} parameter for FGES.
    \item \texttt{blacklist}=$\{\forall{j\in\{1,...,M\}},(Y \rightarrow X^j)\}$, a dataframe format with columns \texttt{from} and \texttt{to} that specifies the forbidden edges.
\end{itemize}

We use all the default parameters for the hybrid MMHC algorithm except for the \texttt{blacklist} parameter that specifies the same structural constraints used for HC. The constraint-based MMPC algorithm within MMHC uses mutual information for conditional independence tests, \texttt{alpha}=$0.05$ that specifies type I error rate for the statistical test, and $max.sx=Inf$ that indicates an unrestricted conditioning set. We use the default parameters for LINGAM.

% \begin{figure}[h]
%     \centering
%     \includegraphics[width=0.5\linewidth]{images/effect_modifier_demo.pdf}
%     \caption{Demonstration of potential effect modifiers. C and D are potential effect modifiers for treatment X and outcome Y.}
%     \label{fig:em_demo}
% \end{figure}
% \subsection*{Potential Effect Modifiers Example}
% In Figure \ref{fig:em_demo}, the nodes $C$ and $D$ are potential effect modifiers for treatment $X$ and outcome $Y$ as these nodes are non-descendants of $X$ and parents of outcome and mediator, respectively.

%% file: images/synthetic_exps/er_logistic_ancEval_support.tex
\begin{table}[!h]
\centering
\begin{adjustbox}{max width=\textwidth}
\begin{tabular}{lrrrrrrrrrrrrrrrr}
\toprule
{Nodes} & \multicolumn{4}{r}{10} & \multicolumn{4}{r}{20} & \multicolumn{4}{r}{30} & \multicolumn{4}{r}{40} \\
{Sparsity} & {0.8} & {1.0} & {1.5} & {2.0} & {0.8} & {1.0} & {1.5} & {2.0} & {0.8} & {1.0} & {1.5} & {2.0} & {0.8} & {1.0} & {1.5} & {2.0} \\
{Support} & {} & {} & {} & {} & {} & {} & {} & {} & {} & {} & {} & {} & {} & {} & {} & {} \\
\midrule
True Causes & {\cellcolor[HTML]{DC5D4A}} \color[HTML]{F1F1F1} 0.88 & {\cellcolor[HTML]{EB7D62}} \color[HTML]{F1F1F1} 0.81 & {\cellcolor[HTML]{E46E56}} \color[HTML]{F1F1F1} 0.84 & {\cellcolor[HTML]{E26952}} \color[HTML]{F1F1F1} 0.86 & {\cellcolor[HTML]{E7745B}} \color[HTML]{F1F1F1} 0.83 & {\cellcolor[HTML]{D24B40}} \color[HTML]{F1F1F1} 0.91 & {\cellcolor[HTML]{F18F71}} \color[HTML]{F1F1F1} 0.78 & {\cellcolor[HTML]{EC7F63}} \color[HTML]{F1F1F1} 0.81 & {\cellcolor[HTML]{CD423B}} \color[HTML]{F1F1F1} 0.93 & {\cellcolor[HTML]{F18F71}} \color[HTML]{F1F1F1} 0.78 & {\cellcolor[HTML]{F18F71}} \color[HTML]{F1F1F1} 0.78 & {\cellcolor[HTML]{E26952}} \color[HTML]{F1F1F1} 0.86 & {\cellcolor[HTML]{CB3E38}} \color[HTML]{F1F1F1} 0.94 & {\cellcolor[HTML]{E67259}} \color[HTML]{F1F1F1} 0.84 & {\cellcolor[HTML]{D75445}} \color[HTML]{F1F1F1} 0.90 & {\cellcolor[HTML]{EE8468}} \color[HTML]{F1F1F1} 0.80 \\
\bottomrule
\end{tabular}
\end{adjustbox}
\caption{Average cause support scores by an ensemble for True and False causes of a sink outcome node for random Erdos-Renyi DAGs with logistic data.}
\end{table}

%% file: images/synthetic_exps/er_logistic_int_ancEval_support.tex
\begin{table}[!h]
\centering
\label{er_logistic_int_ancEval_sup}
\begin{adjustbox}{max width=\textwidth}
\begin{tabular}{lrrrrrrrrrrrrrrrr}
\toprule
{Nodes} & \multicolumn{4}{r}{10} & \multicolumn{4}{r}{20} & \multicolumn{4}{r}{30} & \multicolumn{4}{r}{40} \\
{Sparsity} & {0.8} & {1.0} & {1.5} & {2.0} & {0.8} & {1.0} & {1.5} & {2.0} & {0.8} & {1.0} & {1.5} & {2.0} & {0.8} & {1.0} & {1.5} & {2.0} \\
{Support} & {} & {} & {} & {} & {} & {} & {} & {} & {} & {} & {} & {} & {} & {} & {} & {} \\
\midrule
True Causes & {\cellcolor[HTML]{E16751}} \color[HTML]{F1F1F1} 0.86 & {\cellcolor[HTML]{E46E56}} \color[HTML]{F1F1F1} 0.85 & {\cellcolor[HTML]{EC8165}} \color[HTML]{F1F1F1} 0.81 & {\cellcolor[HTML]{E36B54}} \color[HTML]{F1F1F1} 0.85 & {\cellcolor[HTML]{E97A5F}} \color[HTML]{F1F1F1} 0.82 & {\cellcolor[HTML]{D95847}} \color[HTML]{F1F1F1} 0.89 & {\cellcolor[HTML]{F39577}} \color[HTML]{000000} 0.76 & {\cellcolor[HTML]{E8765C}} \color[HTML]{F1F1F1} 0.83 & {\cellcolor[HTML]{D0473D}} \color[HTML]{F1F1F1} 0.92 & {\cellcolor[HTML]{F18F71}} \color[HTML]{F1F1F1} 0.77 & {\cellcolor[HTML]{EE8468}} \color[HTML]{F1F1F1} 0.80 & {\cellcolor[HTML]{E26952}} \color[HTML]{F1F1F1} 0.86 & {\cellcolor[HTML]{CB3E38}} \color[HTML]{F1F1F1} 0.94 & {\cellcolor[HTML]{E8765C}} \color[HTML]{F1F1F1} 0.83 & {\cellcolor[HTML]{D85646}} \color[HTML]{F1F1F1} 0.89 & {\cellcolor[HTML]{EC8165}} \color[HTML]{F1F1F1} 0.81 \\
\bottomrule
\end{tabular}
\end{adjustbox}
\caption{Average cause support scores by an ensemble for True and False causes of a sink outcome node for random Erdos-Renyi DAGs with logistic+interaction data.}
\end{table}

%% file: images/synthetic_exps/er_linear_ancEval_support.tex
\begin{table}[!h]
\centering
\label{er_logistic_ancEval_sup}
\begin{adjustbox}{max width=\textwidth}
\begin{tabular}{lrrrrrrrrrrrrrrrr}
\toprule
{Nodes} & \multicolumn{4}{r}{10} & \multicolumn{4}{r}{20} & \multicolumn{4}{r}{30} & \multicolumn{4}{r}{40} \\
{Sparsity} & {0.8} & {1.0} & {1.5} & {2.0} & {0.8} & {1.0} & {1.5} & {2.0} & {0.8} & {1.0} & {1.5} & {2.0} & {0.8} & {1.0} & {1.5} & {2.0} \\
{Support} & {} & {} & {} & {} & {} & {} & {} & {} & {} & {} & {} & {} & {} & {} & {} & {} \\
\midrule
True Causes & {\cellcolor[HTML]{DC5D4A}} \color[HTML]{F1F1F1} 0.88 & {\cellcolor[HTML]{EB7D62}} \color[HTML]{F1F1F1} 0.81 & {\cellcolor[HTML]{E46E56}} \color[HTML]{F1F1F1} 0.84 & {\cellcolor[HTML]{E26952}} \color[HTML]{F1F1F1} 0.86 & {\cellcolor[HTML]{E7745B}} \color[HTML]{F1F1F1} 0.83 & {\cellcolor[HTML]{D24B40}} \color[HTML]{F1F1F1} 0.91 & {\cellcolor[HTML]{F18F71}} \color[HTML]{F1F1F1} 0.78 & {\cellcolor[HTML]{EC7F63}} \color[HTML]{F1F1F1} 0.81 & {\cellcolor[HTML]{CD423B}} \color[HTML]{F1F1F1} 0.93 & {\cellcolor[HTML]{F18F71}} \color[HTML]{F1F1F1} 0.78 & {\cellcolor[HTML]{F18F71}} \color[HTML]{F1F1F1} 0.78 & {\cellcolor[HTML]{E26952}} \color[HTML]{F1F1F1} 0.86 & {\cellcolor[HTML]{CB3E38}} \color[HTML]{F1F1F1} 0.94 & {\cellcolor[HTML]{E67259}} \color[HTML]{F1F1F1} 0.84 & {\cellcolor[HTML]{D75445}} \color[HTML]{F1F1F1} 0.90 & {\cellcolor[HTML]{EE8468}} \color[HTML]{F1F1F1} 0.80 \\
\bottomrule
\end{tabular}
\end{adjustbox}
\caption{Average cause support scores by an ensemble for True and False causes of a sink outcome node for random Erdos-Renyi DAGs with logistic data.}
\end{table}

%% file: images/synthetic_exps/ba_logistic_ancEval_support.tex
\begin{table}[!h]
\centering
\label{ba_logistic_ancEval_sup}
\begin{adjustbox}{max width=\textwidth}
\begin{tabular}{lrrrrrrrrrrrrrrrr}
\toprule
{Nodes} & \multicolumn{4}{r}{10} & \multicolumn{4}{r}{20} & \multicolumn{4}{r}{30} & \multicolumn{4}{r}{40} \\
{Sparsity} & {1} & {2} & {3} & {4} & {1} & {2} & {3} & {4} & {1} & {2} & {3} & {4} & {1} & {2} & {3} & {4} \\
{Support} & {} & {} & {} & {} & {} & {} & {} & {} & {} & {} & {} & {} & {} & {} & {} & {} \\
\midrule
True Causes & {\cellcolor[HTML]{EE8669}} \color[HTML]{F1F1F1} 0.79 & {\cellcolor[HTML]{EA7B60}} \color[HTML]{F1F1F1} 0.82 & {\cellcolor[HTML]{F6BFA6}} \color[HTML]{000000} 0.64 & {\cellcolor[HTML]{F7B99E}} \color[HTML]{000000} 0.66 & {\cellcolor[HTML]{EE8669}} \color[HTML]{F1F1F1} 0.79 & {\cellcolor[HTML]{F08A6C}} \color[HTML]{F1F1F1} 0.79 & {\cellcolor[HTML]{F6BEA4}} \color[HTML]{000000} 0.65 & {\cellcolor[HTML]{F59F80}} \color[HTML]{000000} 0.74 & {\cellcolor[HTML]{D95847}} \color[HTML]{F1F1F1} 0.89 & {\cellcolor[HTML]{E67259}} \color[HTML]{F1F1F1} 0.84 & {\cellcolor[HTML]{F59C7D}} \color[HTML]{000000} 0.74 & {\cellcolor[HTML]{F18D6F}} \color[HTML]{F1F1F1} 0.78 & {\cellcolor[HTML]{E36B54}} \color[HTML]{F1F1F1} 0.85 & {\cellcolor[HTML]{F08A6C}} \color[HTML]{F1F1F1} 0.79 & {\cellcolor[HTML]{F18F71}} \color[HTML]{F1F1F1} 0.77 & {\cellcolor[HTML]{F5C2AA}} \color[HTML]{000000} 0.63 \\
\bottomrule
\end{tabular}
\end{adjustbox}
\caption{Average cause support scores by an ensemble for True and False causes of a sink outcome node for random Barabasi-Albert DAGs with logistic data.}
\end{table}

%% file: images/synthetic_exps/ba_logistic_int_ancEval_support.tex
\begin{table}[!h]
\centering
\label{ba_logistic_int_ancEval_sup}
\begin{adjustbox}{max width=\textwidth}
\begin{tabular}{lrrrrrrrrrrrrrrrr}
\toprule
{Nodes} & \multicolumn{4}{r}{10} & \multicolumn{4}{r}{20} & \multicolumn{4}{r}{30} & \multicolumn{4}{r}{40} \\
{Sparsity} & {1} & {2} & {3} & {4} & {1} & {2} & {3} & {4} & {1} & {2} & {3} & {4} & {1} & {2} & {3} & {4} \\
{Support} & {} & {} & {} & {} & {} & {} & {} & {} & {} & {} & {} & {} & {} & {} & {} & {} \\
\midrule
True Causes & {\cellcolor[HTML]{F18F71}} \color[HTML]{F1F1F1} 0.78 & {\cellcolor[HTML]{F29072}} \color[HTML]{F1F1F1} 0.77 & {\cellcolor[HTML]{F7AD90}} \color[HTML]{000000} 0.70 & {\cellcolor[HTML]{F7AC8E}} \color[HTML]{000000} 0.70 & {\cellcolor[HTML]{E97A5F}} \color[HTML]{F1F1F1} 0.82 & {\cellcolor[HTML]{F7AD90}} \color[HTML]{000000} 0.70 & {\cellcolor[HTML]{F6A385}} \color[HTML]{000000} 0.72 & {\cellcolor[HTML]{F6A283}} \color[HTML]{000000} 0.73 & {\cellcolor[HTML]{D24B40}} \color[HTML]{F1F1F1} 0.91 & {\cellcolor[HTML]{DF634E}} \color[HTML]{F1F1F1} 0.87 & {\cellcolor[HTML]{F6A586}} \color[HTML]{000000} 0.72 & {\cellcolor[HTML]{F08B6E}} \color[HTML]{F1F1F1} 0.78 & {\cellcolor[HTML]{E46E56}} \color[HTML]{F1F1F1} 0.85 & {\cellcolor[HTML]{F08A6C}} \color[HTML]{F1F1F1} 0.79 & {\cellcolor[HTML]{E7745B}} \color[HTML]{F1F1F1} 0.83 & {\cellcolor[HTML]{F7B396}} \color[HTML]{000000} 0.68 \\
\bottomrule
\end{tabular}
\end{adjustbox}
\caption{Average cause support scores by an ensemble for True and False causes of a sink outcome node for random Barabasi-Albert DAGs with logistic+interaction data.}
\end{table}

%% file: images/synthetic_exps/ba_linear_ancEval_support.tex
\begin{table}[!h]\label{ba_linear_ancEval_sup}
\centering
\begin{adjustbox}{max width=\textwidth}
\begin{tabular}{lrrrrrrrrrrrrrrrr}
\toprule
{Nodes} & \multicolumn{4}{r}{10} & \multicolumn{4}{r}{20} & \multicolumn{4}{r}{30} & \multicolumn{4}{r}{40} \\
{Sparsity} & {1} & {2} & {3} & {4} & {1} & {2} & {3} & {4} & {1} & {2} & {3} & {4} & {1} & {2} & {3} & {4} \\
{Support} & {} & {} & {} & {} & {} & {} & {} & {} & {} & {} & {} & {} & {} & {} & {} & {} \\
\midrule
True Causes & {\cellcolor[HTML]{E67259}} \color[HTML]{F1F1F1} 0.84 & {\cellcolor[HTML]{F59F80}} \color[HTML]{000000} 0.74 & {\cellcolor[HTML]{F7AF91}} \color[HTML]{000000} 0.69 & {\cellcolor[HTML]{F7B497}} \color[HTML]{000000} 0.68 & {\cellcolor[HTML]{EE8468}} \color[HTML]{F1F1F1} 0.80 & {\cellcolor[HTML]{F7AA8C}} \color[HTML]{000000} 0.71 & {\cellcolor[HTML]{F7B599}} \color[HTML]{000000} 0.67 & {\cellcolor[HTML]{F7B497}} \color[HTML]{000000} 0.68 & {\cellcolor[HTML]{E26952}} \color[HTML]{F1F1F1} 0.86 & {\cellcolor[HTML]{E9785D}} \color[HTML]{F1F1F1} 0.83 & {\cellcolor[HTML]{F7B599}} \color[HTML]{000000} 0.67 & {\cellcolor[HTML]{F7A98B}} \color[HTML]{000000} 0.71 & {\cellcolor[HTML]{EE8468}} \color[HTML]{F1F1F1} 0.80 & {\cellcolor[HTML]{F49A7B}} \color[HTML]{000000} 0.75 & {\cellcolor[HTML]{F18D6F}} \color[HTML]{F1F1F1} 0.78 & {\cellcolor[HTML]{F7B599}} \color[HTML]{000000} 0.67 \\
\bottomrule
\end{tabular}
\end{adjustbox}
\caption{Average cause support scores by an ensemble for True and False causes of a sink outcome node for random Barabasi-Albert DAGs with logistic data.}
\end{table}

%% file: images/synthetic_exps/er_logistic_ancEval.tex
\begin{table*}[!h]
\centering
\label{er_logistic_ancEval}
\begin{adjustbox}{max width=\textwidth}
\begin{tabular}{lllllllllllllllll}
\toprule
{Nodes} & \multicolumn{4}{l}{10} & \multicolumn{4}{l}{20} & \multicolumn{4}{l}{30} & \multicolumn{4}{l}{40} \\
{Sparsity} & {0.8} & {1.0} & {1.5} & {2.0} & {0.8} & {1.0} & {1.5} & {2.0} & {0.8} & {1.0} & {1.5} & {2.0} & {0.8} & {1.0} & {1.5} & {2.0} \\
{Algorithm} & {} & {} & {} & {} & {} & {} & {} & {} & {} & {} & {} & {} & {} & {} & {} & {} \\
\midrule
FGES & {\cellcolor[HTML]{740000}} \color[HTML]{F1F1F1} 0.91±0.08 & {\cellcolor[HTML]{E20000}} \color[HTML]{F1F1F1} 0.82±0.17 & {\cellcolor[HTML]{AB0000}} \color[HTML]{F1F1F1} 0.87±0.12 & {\cellcolor[HTML]{660000}} \color[HTML]{F1F1F1} 0.92±0.14 & {\cellcolor[HTML]{640000}} \color[HTML]{F1F1F1} 0.93±0.04 & {\cellcolor[HTML]{790000}} \color[HTML]{F1F1F1} 0.91±0.13 & {\cellcolor[HTML]{FF6900}} \color[HTML]{F1F1F1} 0.71±0.18 & {\cellcolor[HTML]{FF1200}} \color[HTML]{F1F1F1} 0.78±0.21 & {\cellcolor[HTML]{590000}} \color[HTML]{F1F1F1} 0.93±0.13 & {\cellcolor[HTML]{F20000}} \color[HTML]{F1F1F1} 0.81±0.16 & {\cellcolor[HTML]{FF0800}} \color[HTML]{F1F1F1} 0.79±0.15 & {\cellcolor[HTML]{F90000}} \color[HTML]{F1F1F1} 0.80±0.21 & {\cellcolor[HTML]{100000}} \color[HTML]{F1F1F1} 0.99±0.02 & {\cellcolor[HTML]{C20000}} \color[HTML]{F1F1F1} 0.85±0.16 & {\cellcolor[HTML]{510000}} \color[HTML]{F1F1F1} 0.94±0.09 & {\cellcolor[HTML]{EA0000}} \color[HTML]{F1F1F1} 0.81±0.13 \\
\bottomrule
\end{tabular}
\end{adjustbox}
\caption{F1 scores for discovered causes of a sink outcome node for random Erdos-Renyi DAGs with logistic data.}
\end{table*}

%% file: images/synthetic_exps/er_logistic_int_ancEval.tex
\begin{table*}[!h]
\centering
\label{er_logistic_int_ancEval}
\begin{adjustbox}{max width=\textwidth}
\begin{tabular}{lllllllllllllllll}
\toprule
{Nodes} & \multicolumn{4}{l}{10} & \multicolumn{4}{l}{20} & \multicolumn{4}{l}{30} & \multicolumn{4}{l}{40} \\
{Sparsity} & {0.8} & {1.0} & {1.5} & {2.0} & {0.8} & {1.0} & {1.5} & {2.0} & {0.8} & {1.0} & {1.5} & {2.0} & {0.8} & {1.0} & {1.5} & {2.0} \\
{Algorithm} & {} & {} & {} & {} & {} & {} & {} & {} & {} & {} & {} & {} & {} & {} & {} & {} \\
\midrule
FGES & {\cellcolor[HTML]{BA0000}} \color[HTML]{F1F1F1} 0.91±0.09 & {\cellcolor[HTML]{BD0000}} \color[HTML]{F1F1F1} 0.90±0.12 & {\cellcolor[HTML]{FF4100}} \color[HTML]{F1F1F1} 0.84±0.11 & {\cellcolor[HTML]{8E0000}} \color[HTML]{F1F1F1} 0.93±0.10 & {\cellcolor[HTML]{A00000}} \color[HTML]{F1F1F1} 0.92±0.05 & {\cellcolor[HTML]{C80000}} \color[HTML]{F1F1F1} 0.90±0.14 & {\cellcolor[HTML]{FF6100}} \color[HTML]{F1F1F1} 0.82±0.14 & {\cellcolor[HTML]{FF2C00}} \color[HTML]{F1F1F1} 0.85±0.21 & {\cellcolor[HTML]{8B0000}} \color[HTML]{F1F1F1} 0.93±0.13 & {\cellcolor[HTML]{FF6100}} \color[HTML]{F1F1F1} 0.82±0.15 & {\cellcolor[HTML]{FF1500}} \color[HTML]{F1F1F1} 0.86±0.11 & {\cellcolor[HTML]{E70000}} \color[HTML]{F1F1F1} 0.88±0.12 & {\cellcolor[HTML]{0B0000}} \color[HTML]{F1F1F1} 0.99±0.02 & {\cellcolor[HTML]{FF1700}} \color[HTML]{F1F1F1} 0.86±0.15 & {\cellcolor[HTML]{640000}} \color[HTML]{F1F1F1} 0.95±0.07 & {\cellcolor[HTML]{FF6900}} \color[HTML]{F1F1F1} 0.82±0.17 \\
\bottomrule
\end{tabular}
\end{adjustbox}
\caption{F1 scores for discovered causes of a sink outcome node for random Erdos-Renyi DAGs with logistic data with pairwise interaction.}
\end{table*}

%% file: images/synthetic_exps/er_linear_ancEval.tex
\begin{table*}[!h]
\centering
\label{er_logistic_ancEval}
\begin{adjustbox}{max width=\textwidth}
\begin{tabular}{lllllllllllllllll}
\toprule
{Nodes} & \multicolumn{4}{l}{10} & \multicolumn{4}{l}{20} & \multicolumn{4}{l}{30} & \multicolumn{4}{l}{40} \\
{Sparsity} & {0.8} & {1.0} & {1.5} & {2.0} & {0.8} & {1.0} & {1.5} & {2.0} & {0.8} & {1.0} & {1.5} & {2.0} & {0.8} & {1.0} & {1.5} & {2.0} \\
{Algorithm} & {} & {} & {} & {} & {} & {} & {} & {} & {} & {} & {} & {} & {} & {} & {} & {} \\
\midrule
FGES & {\cellcolor[HTML]{890000}} \color[HTML]{F1F1F1} 0.89±0.10 & {\cellcolor[HTML]{740000}} \color[HTML]{F1F1F1} 0.91±0.09 & {\cellcolor[HTML]{F70000}} \color[HTML]{F1F1F1} 0.80±0.13 & {\cellcolor[HTML]{9E0000}} \color[HTML]{F1F1F1} 0.87±0.14 & {\cellcolor[HTML]{AD0000}} \color[HTML]{F1F1F1} 0.86±0.09 & {\cellcolor[HTML]{FF6E00}} \color[HTML]{F1F1F1} 0.71±0.26 & {\cellcolor[HTML]{FF5400}} \color[HTML]{F1F1F1} 0.73±0.19 & {\cellcolor[HTML]{DF0000}} \color[HTML]{F1F1F1} 0.82±0.17 & {\cellcolor[HTML]{D70000}} \color[HTML]{F1F1F1} 0.83±0.16 & {\cellcolor[HTML]{FF3C00}} \color[HTML]{F1F1F1} 0.75±0.22 & {\cellcolor[HTML]{FF6600}} \color[HTML]{F1F1F1} 0.72±0.17 & {\cellcolor[HTML]{F40000}} \color[HTML]{F1F1F1} 0.81±0.20 & {\cellcolor[HTML]{0B0000}} \color[HTML]{F1F1F1} 0.99±0.04 & {\cellcolor[HTML]{F40000}} \color[HTML]{F1F1F1} 0.80±0.20 & {\cellcolor[HTML]{740000}} \color[HTML]{F1F1F1} 0.91±0.16 & {\cellcolor[HTML]{FF2500}} \color[HTML]{F1F1F1} 0.77±0.14 \\
\bottomrule
\end{tabular}
\end{adjustbox}
\caption{F1 scores for discovered causes of a sink outcome node for random Erdos-Renyi DAGs with linear Bernoulli data.}
\end{table*}

%% file: images/synthetic_exps/ba_logistic_ancEval.tex
\begin{table*}[!h]
\centering
\label{ba_logistic_ancEval}
\begin{adjustbox}{max width=\textwidth}
\begin{tabular}{lllllllllllllllll}
\toprule
{Nodes} & \multicolumn{4}{l}{10} & \multicolumn{4}{l}{20} & \multicolumn{4}{l}{30} & \multicolumn{4}{l}{40} \\
{Sparsity} & {1} & {2} & {3} & {4} & {1} & {2} & {3} & {4} & {1} & {2} & {3} & {4} & {1} & {2} & {3} & {4} \\
{Algorithm} & {} & {} & {} & {} & {} & {} & {} & {} & {} & {} & {} & {} & {} & {} & {} & {} \\
\midrule
FGES & {\cellcolor[HTML]{640000}} \color[HTML]{F1F1F1} 0.87±0.14 & {\cellcolor[HTML]{FC0000}} \color[HTML]{F1F1F1} 0.72±0.14 & {\cellcolor[HTML]{F90000}} \color[HTML]{F1F1F1} 0.72±0.23 & {\cellcolor[HTML]{980000}} \color[HTML]{F1F1F1} 0.82±0.14 & {\cellcolor[HTML]{440000}} \color[HTML]{F1F1F1} 0.90±0.13 & {\cellcolor[HTML]{FF3A00}} \color[HTML]{F1F1F1} 0.65±0.20 & {\cellcolor[HTML]{B80000}} \color[HTML]{F1F1F1} 0.79±0.11 & {\cellcolor[HTML]{FF0A00}} \color[HTML]{F1F1F1} 0.70±0.27 & {\cellcolor[HTML]{440000}} \color[HTML]{F1F1F1} 0.90±0.15 & {\cellcolor[HTML]{B00000}} \color[HTML]{F1F1F1} 0.79±0.20 & {\cellcolor[HTML]{FF5C00}} \color[HTML]{F1F1F1} 0.62±0.18 & {\cellcolor[HTML]{C00000}} \color[HTML]{F1F1F1} 0.78±0.16 & {\cellcolor[HTML]{350000}} \color[HTML]{F1F1F1} 0.92±0.10 & {\cellcolor[HTML]{FF3200}} \color[HTML]{F1F1F1} 0.66±0.22 & {\cellcolor[HTML]{FF5E00}} \color[HTML]{F1F1F1} 0.61±0.29 & {\cellcolor[HTML]{F90000}} \color[HTML]{F1F1F1} 0.72±0.09 \\
\bottomrule
\end{tabular}
\end{adjustbox}
\caption{F1 scores for discovered causes of a sink outcome node for random Barabasi-Albert DAGs with logistic data.}
\end{table*}

%% file: images/synthetic_exps/ba_logistic_int_ancEval.tex
\begin{table*}[!h]
\centering
\label{ba_logistic_int_ancEval}
\begin{adjustbox}{max width=\textwidth}
\begin{tabular}{lllllllllllllllll}
\toprule
{Nodes} & \multicolumn{4}{l}{10} & \multicolumn{4}{l}{20} & \multicolumn{4}{l}{30} & \multicolumn{4}{l}{40} \\
{Sparsity} & {1} & {2} & {3} & {4} & {1} & {2} & {3} & {4} & {1} & {2} & {3} & {4} & {1} & {2} & {3} & {4} \\
{Algorithm} & {} & {} & {} & {} & {} & {} & {} & {} & {} & {} & {} & {} & {} & {} & {} & {} \\
\midrule
FGES & {\cellcolor[HTML]{440000}} \color[HTML]{F1F1F1} 0.89±0.15 & {\cellcolor[HTML]{FF6600}} \color[HTML]{F1F1F1} 0.64±0.15 & {\cellcolor[HTML]{BA0000}} \color[HTML]{F1F1F1} 0.79±0.15 & {\cellcolor[HTML]{710000}} \color[HTML]{F1F1F1} 0.85±0.19 & {\cellcolor[HTML]{4A0000}} \color[HTML]{F1F1F1} 0.89±0.15 & {\cellcolor[HTML]{FF4F00}} \color[HTML]{F1F1F1} 0.66±0.26 & {\cellcolor[HTML]{690000}} \color[HTML]{F1F1F1} 0.86±0.10 & {\cellcolor[HTML]{FF6E00}} \color[HTML]{F1F1F1} 0.63±0.30 & {\cellcolor[HTML]{370000}} \color[HTML]{F1F1F1} 0.90±0.22 & {\cellcolor[HTML]{930000}} \color[HTML]{F1F1F1} 0.82±0.21 & {\cellcolor[HTML]{FF7600}} \color[HTML]{F1F1F1} 0.62±0.22 & {\cellcolor[HTML]{BA0000}} \color[HTML]{F1F1F1} 0.79±0.21 & {\cellcolor[HTML]{180000}} \color[HTML]{F1F1F1} 0.93±0.09 & {\cellcolor[HTML]{FF3F00}} \color[HTML]{F1F1F1} 0.67±0.27 & {\cellcolor[HTML]{FF2200}} \color[HTML]{F1F1F1} 0.70±0.27 & {\cellcolor[HTML]{C20000}} \color[HTML]{F1F1F1} 0.78±0.08 \\
\bottomrule
\end{tabular}
\end{adjustbox}
\caption{F1 scores for discovered causes of a sink outcome node for random Barabasi-Albert DAGs with logistic data with pairwise interactions.}
\end{table*}

%% file: images/synthetic_exps/ba_linear_ancEval.tex
\begin{table*}[!h]
\centering
\label{ba_linear_int_ancEval}
\begin{adjustbox}{max width=\textwidth}
\begin{tabular}{lllllllllllllllll}
\toprule
{Nodes} & \multicolumn{4}{l}{10} & \multicolumn{4}{l}{20} & \multicolumn{4}{l}{30} & \multicolumn{4}{l}{40} \\
{Sparsity} & {1} & {2} & {3} & {4} & {1} & {2} & {3} & {4} & {1} & {2} & {3} & {4} & {1} & {2} & {3} & {4} \\
{Algorithm} & {} & {} & {} & {} & {} & {} & {} & {} & {} & {} & {} & {} & {} & {} & {} & {} \\
\midrule
FGES & {\cellcolor[HTML]{0B0000}} \color[HTML]{F1F1F1} 0.91±0.13 & {\cellcolor[HTML]{EF0000}} \color[HTML]{F1F1F1} 0.74±0.13 & {\cellcolor[HTML]{830000}} \color[HTML]{F1F1F1} 0.82±0.17 & {\cellcolor[HTML]{A80000}} \color[HTML]{F1F1F1} 0.79±0.14 & {\cellcolor[HTML]{D70000}} \color[HTML]{F1F1F1} 0.76±0.26 & {\cellcolor[HTML]{FF9500}} \color[HTML]{000000} 0.62±0.22 & {\cellcolor[HTML]{9B0000}} \color[HTML]{F1F1F1} 0.80±0.07 & {\cellcolor[HTML]{FFB700}} \color[HTML]{000000} 0.59±0.28 & {\cellcolor[HTML]{4C0000}} \color[HTML]{F1F1F1} 0.86±0.24 & {\cellcolor[HTML]{790000}} \color[HTML]{F1F1F1} 0.83±0.17 & {\cellcolor[HTML]{FFAD00}} \color[HTML]{000000} 0.60±0.19 & {\cellcolor[HTML]{890000}} \color[HTML]{F1F1F1} 0.82±0.15 & {\cellcolor[HTML]{0B0000}} \color[HTML]{F1F1F1} 0.91±0.12 & {\cellcolor[HTML]{FF5C00}} \color[HTML]{F1F1F1} 0.66±0.21 & {\cellcolor[HTML]{FF4100}} \color[HTML]{F1F1F1} 0.68±0.27 & {\cellcolor[HTML]{CF0000}} \color[HTML]{F1F1F1} 0.76±0.07 \\
\bottomrule
\end{tabular}
\end{adjustbox}
\caption{F1 scores for discovered causes of a sink outcome node for random Barabasi-Albert DAGs with linear Bernoulli data.}
\end{table*}

%% file: images/synthetic_exps/samples_er_ba_support.tex
\begin{table}[ht]
\centering
\label{samples_ba_linear_ancEval}
\begin{adjustbox}{max width=\textwidth}
\begin{tabular}{lrrrrrrrr}
\toprule
{Topology} & \multicolumn{4}{r}{BA} & \multicolumn{4}{r}{ER} \\
{Samples} & {$n\times25$} & {$n\times50$} & {$n\times100$} & {$n\times500$} & {$n\times25$} & {$n\times50$} & {$n\times100$} & {$n\times500$} \\
{Support} & {} & {} & {} & {} & {} & {} & {} & {} \\
\midrule
True Causes & {\cellcolor[HTML]{EBD3C6}} \color[HTML]{000000} 0.56 & {\cellcolor[HTML]{F4C5AD}} \color[HTML]{000000} 0.62 & {\cellcolor[HTML]{F7BA9F}} \color[HTML]{000000} 0.66 & {\cellcolor[HTML]{F4987A}} \color[HTML]{000000} 0.75 & {\cellcolor[HTML]{CCD9ED}} \color[HTML]{000000} 0.44 & {\cellcolor[HTML]{E4D9D2}} \color[HTML]{000000} 0.53 & {\cellcolor[HTML]{F2CAB5}} \color[HTML]{000000} 0.60 & {\cellcolor[HTML]{F6A385}} \color[HTML]{000000} 0.72 \\
True Causes (sig) & {\cellcolor[HTML]{CEDAEB}} \color[HTML]{000000} 0.45 & {\cellcolor[HTML]{DFDBD9}} \color[HTML]{000000} 0.51 & {\cellcolor[HTML]{E9D5CB}} \color[HTML]{000000} 0.55 & {\cellcolor[HTML]{F6BDA2}} \color[HTML]{000000} 0.65 & {\cellcolor[HTML]{AEC9FC}} \color[HTML]{000000} 0.35 & {\cellcolor[HTML]{C5D6F2}} \color[HTML]{000000} 0.42 & {\cellcolor[HTML]{D7DCE3}} \color[HTML]{000000} 0.48 & {\cellcolor[HTML]{F3C8B2}} \color[HTML]{000000} 0.61 \\
False Causes & {\cellcolor[HTML]{5A78E4}} \color[HTML]{F1F1F1} 0.11 & {\cellcolor[HTML]{6384EB}} \color[HTML]{F1F1F1} 0.13 & {\cellcolor[HTML]{6788EE}} \color[HTML]{F1F1F1} 0.14 & {\cellcolor[HTML]{7093F3}} \color[HTML]{F1F1F1} 0.17 & {\cellcolor[HTML]{516DDB}} \color[HTML]{F1F1F1} 0.07 & {\cellcolor[HTML]{5E7DE7}} \color[HTML]{F1F1F1} 0.12 & {\cellcolor[HTML]{6788EE}} \color[HTML]{F1F1F1} 0.14 & {\cellcolor[HTML]{6A8BEF}} \color[HTML]{F1F1F1} 0.15 \\
False Causes (sig) & {\cellcolor[HTML]{4257C9}} \color[HTML]{F1F1F1} 0.02 & {\cellcolor[HTML]{445ACC}} \color[HTML]{F1F1F1} 0.03 & {\cellcolor[HTML]{465ECF}} \color[HTML]{F1F1F1} 0.04 & {\cellcolor[HTML]{4C66D6}} \color[HTML]{F1F1F1} 0.06 & {\cellcolor[HTML]{3E51C5}} \color[HTML]{F1F1F1} 0.02 & {\cellcolor[HTML]{4055C8}} \color[HTML]{F1F1F1} 0.02 & {\cellcolor[HTML]{445ACC}} \color[HTML]{F1F1F1} 0.03 & {\cellcolor[HTML]{4961D2}} \color[HTML]{F1F1F1} 0.05 \\
\bottomrule
\end{tabular}
\end{adjustbox}
\caption{Sensitivity of average cause support to potential unfaithfulness due to limited sample size for $n=30$, and $sp_{ba}=sp_{er}=2$. For limited samples, CSL algorithms may miss true causes as well. A ranking scheme utilizing cause support, magnitude of causal effects, and significance support may be more favorable than a voting ensemble.}
\end{table}